\documentclass[journal]{IEEEtran}

\usepackage{url}            
\usepackage{booktabs}       
\usepackage{amsfonts}       
\usepackage{nicefrac}       
\usepackage{microtype}      
\usepackage{xcolor}         
\usepackage{amsmath}
\usepackage{amssymb}
\usepackage{mathtools}
\usepackage{amsthm} 
\usepackage{enumitem}
\usepackage{bm}
\usepackage{multirow}
\usepackage{graphicx}
\usepackage{wrapfig}
\usepackage{subcaption} 
\usepackage[table]{xcolor}
\usepackage{xspace}
\usepackage{pifont}
\usepackage{dblfloatfix}
\usepackage{hyperref}
\newcommand{\model}{SUPRA\xspace}

\newtheorem{theorem}{Theorem}

\newtheorem{proposition}{Proposition}
\newtheorem{corollary}[theorem]{Corollary}
\newtheorem{definition}[theorem]{Definition}

\title{Beyond the Aggregation Dilemma: Prior-Retaining Decoupled Learning for Multimodal Graphs}

\author{
\IEEEauthorblockN{
Hao Yan\IEEEauthorrefmark{1},
Xuanru Wang\IEEEauthorrefmark{1},
Jun Yin,
Shirui Pan,~\IEEEmembership{Senior Member,~IEEE},\\
Senzhang Wang\IEEEauthorrefmark{2},~\IEEEmembership{Member,~IEEE} 
and Chengqi Zhang\IEEEauthorrefmark{2},~\IEEEmembership{Fellow,~IEEE} 
}
\IEEEcompsocitemizethanks{
\IEEEcompsocthanksitem Hao Yan is with the School of Computer Science and Engineering, Central South University, Changsha, China, and also with the Department of Data Science and Artificial Intelligence, The Hong Kong Polytechnic University, Hong Kong SAR, China. This work was conducted during his visit to The Hong Kong Polytechnic University.
Email: CSUyh1999@csu.edu.cn, hao99.yan@polyu.edu.hk
\IEEEcompsocthanksitem Xuanru Wang and Senzhang Wang are with the School of Computer Science and Engineering, Central South University, Changsha, China. 
Email: wangtiant@csu.edu.cn, szwang@csu.edu.cn  
\IEEEcompsocthanksitem Jun Yin and Chengqi Zhang are with the Department of Data Science and Artificial Intelligence, The Hong Kong Polytechnic University, Hong Kong SAR, China. 
Email: Junmay.yin@connect.polyu.hk, Chengqi.zhang@polyu.edu.hk
\IEEEcompsocthanksitem Shirui Pan is with the School of Information and Communication Technology, Griffith University, Brisbane, Australia. 
Email: s.pan@griffith.edu.au
}
\thanks{* Equal Contribution. \dag~Corresponding Authors.}
}

\begin{document}

\maketitle

\begin{abstract}

Multimodal Attributed Graph Learning (MAGL) integrates intrinsic node attributes with structural topology via graph aggregation. 
However, as pretrained encoders evolve into Large Foundation Models 
(LFMs), the landscape of MAGL fundamentally shifts: under high-confidence LFM priors, mandatory aggregation introduces topological noise that overwhelms discriminative signals, triggering a counter-intuitive \textit{performance inversion} where sophisticated MAGL architectures underperform simple topology-agnostic MLPs.
Through systematic empirical and theoretical analysis, we identify that this inversion stems from a fundamental {aggregation dilemma} characterized by two concurrent pathologies: (1) Representational Pathology (SNR Degradation)—mandatory aggregation dilutes robust intrinsic features with topological noise, causing the noise penalty to outweigh its collaborative benefit; and (2) {Optimization Pathology (Gradient Starvation)}—topological aggregation attenuates gradient flow, while a shared task loss causes dominant modalities to prematurely suppress weaker ones. To resolve this dilemma, we propose {\model} (Shared-Unique Prior-Retaining Architecture), a decoupled dual-pathway paradigm. SUPRA processes modality-specific features through topology-agnostic MLPs while capturing structural synergy via a lightweight shared GNN, with auxiliary deep supervision counteracting gradient starvation. Extensive evaluations demonstrate that SUPRA achieves state-of-the-art performance while requiring 3.5$\times$ lower peak GPU memory and up to 4.4$\times$ faster training time than Multimodal Graph Transformers.
\end{abstract}

\begin{IEEEkeywords}
Multimodal Attributed Graphs, Large Foundation Models, Graph Neural Networks.
\end{IEEEkeywords}

\section{Introduction}\label{sec:intro}

Graphs are ubiquitous in the real world, where nodes correspond to entities and edges capture their relationships~\cite{GCN,GAT}. In many practical scenarios~\cite{ed2,netflix,Transformers4Rec}, nodes are associated with multimodal attributes (e.g., text and images), giving rise to Multimodal Attributed Graphs (MAGs)~\cite{yan2025graph, peng2024learning}.
Effective learning on MAGs requires characterizing nodes from two complementary perspectives: the node-level intrinsic multimodal attributes and the structural context provided by the graph topology. A fundamental challenge lies in integrating these information sources.

To address this challenge, Multimodal Attributed Graph Learning (MAGL) paradigms have evolved from early Multimodal GNNs (e.g., MMGCN~\cite{mmgcn} and MGAT~\cite{mgat}) to advanced Multimodal Graph Transformers (e.g., MIG-GT~\cite{MIGGT}, NTSFormer~\cite{ntsformer}). 
Historically, to compensate for the limited expressiveness of initial node representations, these architectures heavily relied on complex aggregation mechanisms to deeply fuse intrinsic semantics with structural topology~\cite{MAGSurvery}.
Concurrently, the rapid advancement of Foundation Models (FMs) has transformed representation learning~\cite{llm_semantic,Exploring,CS-TAG}. As feature extractors evolve from classic specialized FMs (e.g., RoBERTa~\cite{Roberta} and CLIP~\cite{CLIP}) to advanced Large Foundation Models (LFMs, e.g., Llama-3.2-11B-Vision~\cite{llamav2}), the quality of initial node attributes fundamentally shifts from noisy inputs to highly discriminative semantic priors.
This paradigm shift raises a critical question: \textit{in the era of powerful Foundation Models, is structural context still necessary for multimodal graphs, and if so, can existing MAGL architectures effectively integrate topology with strong semantic priors?}

\begin{figure*}[t]
    \centering
    \begin{subfigure}[b]{0.52\textwidth}
        \centering
        \includegraphics[width=\linewidth, height=5.5cm, keepaspectratio]{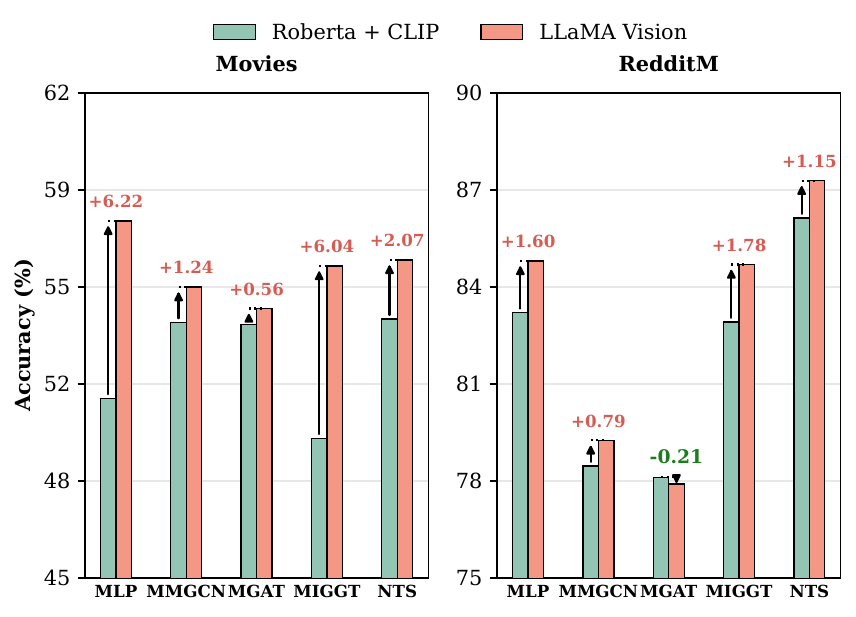}
        \caption{Overall Performance Comparison.}
        \label{fig:intro_perf} 
    \end{subfigure}
    \hfill
    \begin{subfigure}[b]{0.46\textwidth} 
        \centering
        \includegraphics[width=\linewidth, height=5.5cm, keepaspectratio]{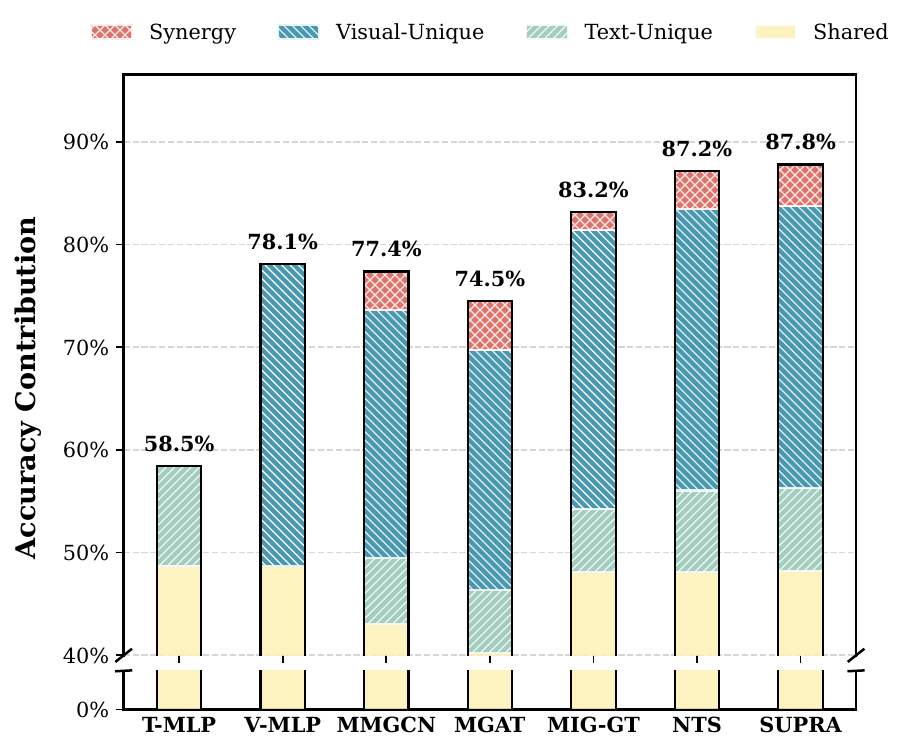}
        \caption{Semantic Attribution Analysis.}
        \label{fig:intro_attr} 
    \end{subfigure}
    
    \caption{\textbf{Performance inversion and semantic attribution.} \textbf{(a)} Accuracy comparison under classic FMs versus high-confidence LFMs, demonstrating the unexpected superiority of topology-agnostic MLPs. \textbf{(b)} Semantic attribution on Reddit-M isolating intrinsic node features (\textit{Unique}/\textit{Shared}) from structural context (\textit{Synergy}).}
    \vspace{-0.5cm}
    \label{fig:main_intro} 
\end{figure*}

Motivated by this question, we conduct a pilot study comparing a topology-agnostic Multimodal MLP against representative MAGL models under the two feature regimes (Figure~\ref{fig:main_intro}a). Under the classic FM regime (RoBERTa+CLIP), structural context is clearly beneficial, with MAGL models generally outperforming the MLP. However, upgrading to LLaMA-Vision features triggers a counter-intuitive \textit{performance inversion}. While most models improve, existing graph architectures fail to meaningfully capitalize on topology. Specifically, the topology-free MLP surpasses traditional multimodal GNNs across the board, notably achieving the highest overall accuracy on the Movies dataset. On Reddit-M, only the state-of-the-art NTSFormer manages to outperform the MLP, but it relies on a heavily parameterized architecture that incurs a prohibitive computational cost. This indicates that complex aggregation struggles to justify its severe overhead under high-confidence priors.

To uncover the root cause, our semantic attribution analysis (Figure~\ref{fig:main_intro}b) explicitly breaks down the prediction accuracy into two fundamental sources: the baseline performance derived from the node's own intrinsic features (comprising modality-\textit{Unique} and \textit{Shared} semantics), and the collaborative gain extracted from the graph topology (\textit{Synergy}).
The results reveal that existing paradigms are trapped in a fundamental {\textit{aggregation dilemma}}: while traditional Multimodal GNNs employ rigid aggregation that dilutes robust intrinsic features with topological noise, Graph Transformers such as MIG-GT better preserve these features but employ weak aggregation that starves cross-modal synergy. 
As semantic priors become dominant, the noise penalty of aggregation begins to outweigh its collaborative benefit, turning this inherent dilemma into a critical bottleneck. Furthermore, while state-of-the-art models like NTSFormer attempt to bypass this by feeding explicitly concatenated multi-hop features into a Transformer, this brute-force dimension expansion incurs prohibitive overhead and still fails to escape the intrinsic topological noise, posing a severe risk of overfitting, notably failing to outperform the topology-agnostic MLP on the Movies dataset (Figure~\ref{fig:main_intro}a).

We theoretically formalize this architectural dilemma by analyzing learning dynamics through the lens of Signal-to-Noise Ratio (SNR)~\cite{Revisiting,SGC} and optimization pathology (\S\ref{sec:theory}). We demonstrate that mandatory topological aggregation intrinsically degrades the SNR of high-confidence semantic priors (\S\ref{sec:theory_snr}). Furthermore, we identify a dual-bottleneck optimization pathology termed \textit{Gradient Starvation}~\cite{Gradient,OGMGE}, where topological aggregation attenuates gradient flow, and a shared task loss allows dominant modalities to prematurely starve weaker ones (\S\ref{sec:theory_grad}).
Guided by these theoretical insights, we propose \textbf{\model} (\textbf{S}hared-\textbf{U}nique \textbf{P}rior-\textbf{R}etaining \textbf{A}rchitecture), a straightforward yet effective decoupled paradigm for multimodal graph learning. Diverging from the trend of increasing architectural complexity, SUPRA advocates {selective propagation} to explicitly model the two complementary perspectives of MAGs defined at the outset: {node-level intrinsic multimodal attributes} and {structural context}. Specifically, \model establishes a {decoupled dual-pathway architecture}: it employs topology-agnostic MLPs as modality projectors to process and compress \textit{modality-unique} features into \textit{specificity streams}, which strictly bypass the GNN to preserve high-confidence {multimodal attributes}.
Concurrently, a lightweight \textit{synergy stream} (e.g., employing a foundational GNN encoder~\cite{GCN}) jointly processes these features to capture {structural context} by extracting structural consensus. 
To counteract the {Gradient Starvation} triggered by topology-diluted gradients, we apply auxiliary deep supervision directly to the specificity branches, ensuring they receive clean, topology-agnostic signals.
Ultimately, extensive experiments demonstrate that this simplified architecture achieves state-of-the-art performance with remarkably low computational overhead. Our main contributions are summarized as follows:

\begin{enumerate}[leftmargin=*,noitemsep]
\item We uncover a counter-intuitive \textit{performance inversion}: under high-confidence LFM features, basic topology-agnostic MLPs match or exceed sophisticated MAGL models. Through semantic attribution, we identify the root cause as a fundamental {aggregation dilemma}: mandatory aggregation dilutes robust intrinsic semantics with topological noise—a bottleneck that becomes increasingly pronounced as feature quality improves.

\item 
We theoretically analyze this aggregation dilemma from two perspectives. First, we prove that mean-aggregation {SNR degradation}—under high-confidence priors, the post-aggregation SNR is strictly bounded below the intrinsic SNR. Second, we identify {Gradient Starvation}—a shared task loss causes dominant modalities to prematurely suppress weaker ones during joint training. 
These two mechanisms formalize why coupled MAGL architectures cannot simultaneously preserve semantic quality and capture structural synergy.

\item  We propose \model, a decoupled dual-pathway architecture that elegantly resolves this {aggregation dilemma} through topology-agnostic specificity streams and a lightweight synergy stream.
By counteracting gradient starvation via auxiliary supervision, \model achieves state-of-the-art performance with 3.5× lower peak GPU memory and 4.4× faster training than Multimodal Graph Transformers, demonstrating superior scalability.



\end{enumerate}

\section{Preliminaries}
\label{sec:pre}

\subsection{Multimodal Attributed Graph \& Problem Formulation}
A Multimodal Attributed Graph (MAG) is denoted as $\mathcal{G} = (\mathcal{V}, \mathcal{E}, \mathcal{X})$, where $\mathcal{V}$ is the set of $N$ nodes and $\mathcal{E}$ is represented by an adjacency matrix $\mathbf{A} \in \{0, 1\}^{N \times N}$. Each node $v \in \mathcal{V}$ is associated with multimodal features $\mathcal{X}_v = \{ \mathbf{x}_v^{(m)} \in \mathbb{R}^{d_m} \}_{m \in \mathcal{M}}$. Extracted via Large Foundation Models (e.g., LLaMA-Vision~\cite{llamav2}), these features serve as \textit{High-Confidence Semantic Priors} with rich intrinsic knowledge. Building upon $\mathcal{G}$, we focus on node classification. Given a labeled subset $\mathcal{V}_{train}$, the goal is to learn a mapping $F_{\theta}: \left( \mathbf{A}, \{ \mathbf{X}^{(m)} \}_{m \in \mathcal{M}} \right) \to \mathbf{Z} \in \mathbb{R}^{N \times C}$ that minimizes the prediction error on $\mathcal{V}_{train}$ to accurately predict probabilities over $C$ classes for unlabeled nodes.

\subsection{Message-Passing Paradigms on MAGs}
\label{sec:mp_framework}

Modern GNNs follow a message-passing framework~\cite{gilmer2017neural}, iteratively updating node embeddings as:
\begin{equation}
    \mathbf{h}_v^{(\ell+1)} = \mathrm{UPDATE}\Big( \mathbf{h}_v^{(\ell)},\ \mathrm{AGG}_{u \in \mathcal{N}_v} \big( \mathrm{MSG}(\mathbf{h}_u^{(\ell)}) \big) \Big).
\end{equation}
To enable a rigorous analysis of how intrinsic node semantics interact with graph structure, we focus on the canonical {mean-aggregation} operator:
\begin{equation}
\label{eq:mean_aggr}
    \mathbf{h}_v^{(\ell+1)} = \alpha \mathbf{h}_v^{(\ell)} + (1-\alpha) \frac{1}{|\mathcal{N}_v|} \sum_{u \in \mathcal{N}_v} \mathbf{h}_u^{(\ell)},
\end{equation}
where $\alpha \in (0,1)$ is the self-retention weight. Based on how multimodal features $\mathcal{X}_v = \{ \mathbf{x}_v^{(m)} \}_{m \in \mathcal{M}}$ are coupled with this operator, existing methods fundamentally fall into two paradigms:

\noindent\textbf{(1) Joint Aggregation (Standard GNNs).} 
Modalities are fused \emph{prior} to propagation, entangling all modalities through a single shared smoothing operator:
\begin{equation}
    \mathbf{x}_v^{\text{joint}} = \text{Concat}(\{ \mathbf{x}_v^{(m)} \}), \quad
    \mathbf{h}_v = \text{GNN}_{\text{aggr}}(\mathcal{N}_v, \mathbf{x}_v^{\text{joint}}).
\end{equation}

\noindent\textbf{(2) Independent Aggregation (Multimodal GNNs).} 
Modality-specific structures are preserved by decoupling propagation, deferring cross-modal interaction to a post-hoc fusion stage:
\begin{equation}
    \mathbf{h}_v^{(m)} = \text{GNN}_{\text{aggr}}(\mathcal{N}_v, \mathbf{x}_v^{(m)}), \quad
    \mathbf{z}_v = \text{Fusion}(\{ \mathbf{h}_v^{(m)} \}).
\end{equation}


\begin{figure*}[!t]
    \centering
    \includegraphics[width=\linewidth]{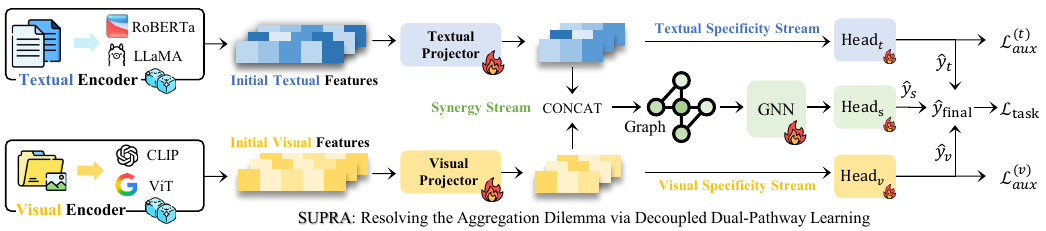}
    \caption{Overview of \model: Topology-agnostic specificity streams and a lightweight synergy stream, resolved via auxiliary deep supervision.}
    \label{fig:supra}
\end{figure*}

\section{The Aggregation Dilemma: Theory and Analysis}
\label{sec:theory}

We analyze why standard message passing fails in the presence of \textit{High-Confidence Semantic Priors}. Our key observation is that mean-aggregation, the canonical operation underlying most GNNs~\cite{GCN,SAGE,SGC}, introduces \emph{topological noise} that can overwhelm already strong semantic signals, and further induces an optimization bias that suppresses weaker modalities.

\subsection{Representational Pathology: SNR Degradation}
\label{sec:theory_snr}

Formally, for node $v$ and modality $m$, we decompose the feature as $\mathbf{x}_v^{(m)} = \mathbf{s}_v^{(m)} + \boldsymbol{\epsilon}_v^{(m)}$, where $\mathbf{s}_v^{(m)}$ is the true semantic signal and $\boldsymbol{\epsilon}_v^{(m)}$ is encoder noise with variance $\sigma_\epsilon^2$. Here $\alpha \in (0,1)$ denotes the self-loop retention weight in mean-aggregation (Eq~\ref{eq:mean_aggr}), $\beta \in [0,1]$ denotes the semantic alignment coefficient, and $\sigma_{\mathcal{N}}^2$ denotes the variance of the induced neighborhood noise.

\begin{theorem}[SNR Degradation under Mean-Aggregation]
\label{thm:snr}
Consider mean-aggregation (Eq.~\ref{eq:mean_aggr}). There exists a critical noise threshold:
\begin{equation}
\label{eq:tau_main}
    \tau(\beta) = \frac{1-\alpha}{\beta\bigl(2\alpha + (1-\alpha)\beta\bigr)} \, \sigma_{\mathcal{N}}^2,
\end{equation}
such that the post-aggregation Signal-to-Noise Ratio (SNR) is strictly degraded ($\text{SNR}_{post}^{(m)} < \text{SNR}_{int}^{(m)}$) if and only if the intrinsic encoder noise satisfies $\sigma_\epsilon^2 < \tau(\beta)$.
\end{theorem}

\noindent
\textbf{Implication \& Empirical Validation.} The condition $\sigma_\epsilon^2 < \tau(\beta)$ defines the regime of \textit{High-Confidence Semantic Priors}. This formally establishes that when features are already high quality, aggregation is guaranteed to reduce representational quality—rather than a hyperparameter-tuning failure, this is a structural inevitability of mean aggregation. 
Our controlled noise-injection experiments (\S\ref{sec:exp_noise}) directly observe this crossover: as feature noise decreases, MLPs systematically outperform aggregation-based models, consistent with the theoretical prediction. See Appendix~\ref{app:proof_thm1} for the complete proof.

\subsection{Optimization Pathology: Gradient Starvation}
\label{sec:theory_grad}
Beyond the representational failure identified in Theorem~\ref{thm:snr}, aggregation architectures also distort optimization dynamics.
In multimodal settings with modality asymmetry (e.g., strong text vs. weak visual, where $\text{SNR}^{(t)} \gg \text{SNR}^{(v)}$), we identify a systematic suppression of weaker modalities governed by two compounding bottlenecks.

\begin{proposition}[Gradient Starvation under Modality Asymmetry]
\label{prop:starvation}
Consider a multi-branch architecture where a weaker modality encoder $f_v$ is optimized via a shared task loss $\mathcal{L}_{\text{task}} = \sum_i \mathcal{L}_i$. For an arbitrary node $i$, the expected gradient norm flowing to $f_v$ is bounded by:
\begin{equation}
\label{eq:grad_bound}
    \mathbb{E}\left[\|\nabla_{f_v}\mathcal{L}_i\|\right] \;\leq\; \underbrace{\eta}_{\text{Architectural}} \cdot \underbrace{|r_i(t)| \cdot C}_{\text{Competition}},
\end{equation}
where $\eta = \alpha^L$ if routed through $L$ GNN layers and $\eta = 1$ for a topology-agnostic bypass, $r_i(t) = \hat{y}_i(t) - y_i$ denotes the joint training residual at step $t$, and $C = \|\mathbf{w}_v\| \cdot \left\|\frac{\partial \mathbf{h}_i^{(v)}}{\partial f_v}\right\|$ is a bounded constant independent of training dynamics.
\end{proposition}

\noindent
\textbf{Implication \& Empirical Validation.} Proposition~\ref{prop:starvation} predicts two concurrent failure modes. First, \textit{Architectural Bottleneck}: forcing features through $L$ GNN layers attenuates the expected gradient flow by $\mathcal{O}(\alpha^L)$. Second, \textit{Optimization Bottleneck}: even with a topology-agnostic bypass ($\eta = 1$), a shared prediction head causes the dominant modality to suppress the weaker one as the joint residual $r_i(t) \to 0$. We verify both mechanisms through controlled experiments: (1) ablating the bypass pathway isolates the architectural effect (see \S\ref{sec:exp_grad}), and (2) adding independent per-modality supervision directly counteracts the optimization bottleneck (see \S\ref{sec:exp_grad}). See Appendix~\ref{app:proof_thm2} for the complete proof.


\subsection{Resolving the Aggregation Dilemma}
\label{sec:theory_dilemma}

Theorems~\ref{thm:snr} and Proposition~\ref{prop:starvation} reveal a fundamental {Aggregation Dilemma}: mandatory topological aggregation simultaneously (i) degrades strong semantic priors via SNR reduction, and (ii) suppresses weaker modalities via gradient starvation. 
Resolving this dilemma requires satisfying two necessary architectural conditions: (1) \textit{Topological Decoupling}---modality-specific intrinsic semantics must bypass aggregation to preserve their intrinsic SNR; and (2) \textit{Gradient Isolation}---all modalities require independent supervision pathways to escape domination by the shared gradient flow.
In \S\ref{sec:methods}, we present \model, whose decoupled dual-pathway architecture naturally satisfies both conditions.

\section{The Proposed Paradigm: \model}
\label{sec:methods}

As established in Section~\ref{sec:theory}, resolving the aggregation dilemma requires fundamentally decoupling intrinsic semantics from graph topology. Rather than adopting computationally expensive Multimodal Graph Transformers~\cite{MIGGT,ntsformer} that implicitly mix these signals, we propose \textbf{\model} (\textbf{S}hared-\textbf{U}nique \textbf{P}rior-\textbf{R}etaining \textbf{A}rchitecture). \model efficiently counteracts both SNR degradation and gradient starvation through a straightforward, decoupled dual-pathway architecture.

\subsection{Decoupled Dual-Pathway Architecture}
\label{sec:architecture}
\model processes multimodal features $\mathbf{X}^{(m)}$ through two distinct functional pathways.

\noindent\textbf{Unique Specificity Streams}.
To satisfy the first necessary condition and prevent SNR degradation, \model strictly isolates modality-specific semantics from topological noise. For each modality $m \in \mathcal{M}$, we employ a base projector (parameterized as an MLP, $f_m$) to process the input features:
\begin{equation}
\label{eq:unique}
    \mathbf{Z}^{(U_m)} = f_m\!\left(\mathbf{X}^{(m)}\right), \qquad \forall m\in\mathcal{M}.
\end{equation}
By bypassing the GNN entirely, these streams act as topology-agnostic pathways, perfectly preserving the intrinsic SNR of the foundation model priors.

\noindent\textbf{Synergy Stream.}
To capture cross-modal topological synergy, we concatenate the \textit{projected} representations and feed them into a foundational GNN (e.g., GCN~\cite{GCN}) for structural propagation:
\begin{equation}
\label{eq:shared}
    \mathbf{H}^{(S)} = \mathrm{Concat}\left(\{\mathbf{Z}^{(U_m)}\}_{m\in\mathcal{M}}\right), \mathbf{Z}^{(S)} = \mathrm{GNN}\left(\mathbf{A}, \mathbf{H}^{(S)}\right).
\end{equation}
Note that we propagate the lower-dimensional projections $\mathbf{Z}^{(U_m)}$ rather than the raw features $\mathbf{X}^{(m)}$, significantly reducing the computational overhead of the message-passing phase.

\noindent\textbf{Parallel Prediction Heads.}
The modality-specific projectors $f_m$ produce topology-agnostic representations $\mathbf{Z}^{(U_m)}$, which are classified by dedicated linear heads $\text{Head}_m$. The final joint prediction is obtained via mean pooling of all streams:
\begin{equation}
    \hat{\mathbf{y}}_{\text{final}} = \frac{1}{|\mathcal{M}|+1}\Big(\text{Head}_S(\mathbf{Z}^{(S)}) + \sum_{m \in \mathcal{M}} \text{Head}_m(\mathbf{Z}^{(U_m)})\Big).
\end{equation}

\subsection{Optimization Objective}
\label{sec:objectives}

The primary task loss supervises the final fused prediction via the standard cross-entropy loss:
\begin{equation}
    \mathcal{L}_{\text{task}} = \mathcal{L}_{\text{CE}}(\hat{\mathbf{y}}_{\text{final}}, \mathbf{y}).
\end{equation}

\noindent\textbf{Auxiliary Deep Supervision.}
However, $\mathcal{L}_{\text{task}}$ alone cannot reliably overcome {Gradient Starvation}: when all streams share a single prediction head, the Synergy Stream may rapidly dominate the gradient flow, effectively suppressing the Unique Streams. To satisfy this necessary condition, we apply auxiliary deep supervision directly to each Unique Stream:
\begin{equation}
    \mathcal{L}_{\text{aux}} = \sum_{m \in \mathcal{M}} \mathcal{L}_{\text{aux}}^{(m)} = \sum_{m \in \mathcal{M}} \mathcal{L}_{\text{CE}}\!\left(\text{Head}_m(\mathbf{Z}^{(U_m)}), \mathbf{y}\right).
\end{equation}
This provides each modality projector with an independent gradient pathway that is immune to domination by the Synergy Stream, ensuring robust training for all modalities regardless of their relative strength. The total training objective is:
\begin{equation}
    \mathcal{L}_{\text{total}} = \mathcal{L}_{\text{task}} + \lambda_{\text{aux}} \, \mathcal{L}_{\text{aux}},
\end{equation}
where $\lambda_{\text{aux}} > 0$ controls the strength of the auxiliary supervision.

\section{Experiments}
\label{sec:exp}

\definecolor{sagegreen}{RGB}{162, 211, 180}

\newcommand{\first}[1]{\cellcolor{sagegreen}\textbf{#1}}
\newcommand{\second}[1]{\cellcolor{sagegreen!65}\underline{#1}}
\newcommand{\third}[1]{\cellcolor{sagegreen!30}#1}

\begin{table*}[t]
\centering
\caption{\textbf{Overall Performance Comparison.} We evaluate SUPRA against baselines across four datasets. The \colorbox{sagegreen}{\textbf{best}}, \colorbox{sagegreen!65}{\underline{second-best}}, and \colorbox{sagegreen!30}{third-best} results are highlighted.}
\label{tab:main_perf}
\resizebox{\textwidth}{!}{
\renewcommand{\arraystretch}{1.15} 
\begin{tabular}{l l cccccccc}
\toprule
\multirow{2.5}{*}{\textbf{Backbone}} & \multirow{2.5}{*}{\textbf{Method}} & \multicolumn{2}{c}{\textbf{Movies}} & \multicolumn{2}{c}{\textbf{Grocery}} & \multicolumn{2}{c}{\textbf{Toys}} & \multicolumn{2}{c}{\textbf{Reddit-M}} \\
\cmidrule(lr){3-4} \cmidrule(lr){5-6} \cmidrule(lr){7-8} \cmidrule(lr){9-10}
& & Acc & F1 & Acc & F1 & Acc & F1 & Acc & F1 \\ \midrule

\rowcolor{gray!10} 
\multicolumn{10}{c}{\textit{Feature Group: Standard Priors (CLIP+RoBERTa)}} \\ \midrule

\multirow{3}{*}{\textbf{MLP}} 
  & Text   & 46.52 \scalebox{0.7}{$\pm$ 0.40} & 34.25 \scalebox{0.7}{$\pm$ 0.26} & 78.02 \scalebox{0.7}{$\pm$ 0.36} & 68.95 \scalebox{0.7}{$\pm$ 0.33} & 73.89 \scalebox{0.7}{$\pm$ 0.22} & 71.50 \scalebox{0.7}{$\pm$ 0.30} & 46.98 \scalebox{0.7}{$\pm$ 0.18} & 44.78 \scalebox{0.7}{$\pm$ 0.23} \\
  & Visual & 49.06 \scalebox{0.7}{$\pm$ 0.05} & 38.09 \scalebox{0.7}{$\pm$ 0.61} & 67.86 \scalebox{0.7}{$\pm$ 0.39} & 59.15 \scalebox{0.7}{$\pm$ 0.69} & 68.90 \scalebox{0.7}{$\pm$ 0.26} & 65.79 \scalebox{0.7}{$\pm$ 0.51} & 78.27 \scalebox{0.7}{$\pm$ 0.04} & 73.65 \scalebox{0.7}{$\pm$ 0.08} \\
  & EF-MLP & 51.29 \scalebox{0.7}{$\pm$ 0.40} & 41.40 \scalebox{0.7}{$\pm$ 0.33} & 80.62 \scalebox{0.7}{$\pm$ 0.46} & 72.82 \scalebox{0.7}{$\pm$ 0.17} & 76.73 \scalebox{0.7}{$\pm$ 0.01} & 74.14 \scalebox{0.7}{$\pm$ 0.10} & 83.20 \scalebox{0.7}{$\pm$ 0.10} & 79.78 \scalebox{0.7}{$\pm$ 0.09} \\ \cmidrule{1-10}

\multirow{4}{*}{\textbf{GNNs}} 
  & GCN    & 54.41 \scalebox{0.7}{$\pm$ 0.13} & 46.56 \scalebox{0.7}{$\pm$ 0.68} & 82.25 \scalebox{0.7}{$\pm$ 0.06} & 72.09 \scalebox{0.7}{$\pm$ 0.22} & 80.08 \scalebox{0.7}{$\pm$ 0.16} & 77.48 \scalebox{0.7}{$\pm$ 0.15} & 78.42 \scalebox{0.7}{$\pm$ 0.05} & 74.39 \scalebox{0.7}{$\pm$ 0.17} \\
  & GAT    & 53.88 \scalebox{0.7}{$\pm$ 0.27} & 46.93 \scalebox{0.7}{$\pm$ 1.28} & 81.74 \scalebox{0.7}{$\pm$ 0.01} & 71.72 \scalebox{0.7}{$\pm$ 0.49} & 80.22 \scalebox{0.7}{$\pm$ 0.12} & 77.62 \scalebox{0.7}{$\pm$ 0.08} & 77.96 \scalebox{0.7}{$\pm$ 0.12} & 73.86 \scalebox{0.7}{$\pm$ 0.07} \\
  & SAGE   & \first{56.03 \scalebox{0.7}{$\pm$ 0.32}} & \second{49.66 \scalebox{0.7}{$\pm$ 0.95}} & 83.29 \scalebox{0.7}{$\pm$ 0.04} & \third{74.79 \scalebox{0.7}{$\pm$ 0.53}} & \second{80.64 \scalebox{0.7}{$\pm$ 0.23}} & \second{78.18 \scalebox{0.7}{$\pm$ 0.29}} & \third{86.44 \scalebox{0.7}{$\pm$ 0.08}} & \third{83.71 \scalebox{0.7}{$\pm$ 0.05}} \\
  & JKNet  & 54.41 \scalebox{0.7}{$\pm$ 0.19} & 48.35 \scalebox{0.7}{$\pm$ 0.82} & 82.74 \scalebox{0.7}{$\pm$ 0.22} & 72.38 \scalebox{0.7}{$\pm$ 0.49} & 80.27 \scalebox{0.7}{$\pm$ 0.20} & 77.60 \scalebox{0.7}{$\pm$ 0.06} & 78.19 \scalebox{0.7}{$\pm$ 0.05} & 74.06 \scalebox{0.7}{$\pm$ 0.17} \\ \cmidrule{1-10}

\multirow{4}{*}{\textbf{MAGs}} 
  & MMGCN  & 53.95 \scalebox{0.7}{$\pm$ 0.34} & 46.08 \scalebox{0.7}{$\pm$ 0.46} & 81.53 \scalebox{0.7}{$\pm$ 0.38} & 72.09 \scalebox{0.7}{$\pm$ 0.16} & 80.05 \scalebox{0.7}{$\pm$ 0.08} & 77.19 \scalebox{0.7}{$\pm$ 0.25} & 78.46 \scalebox{0.7}{$\pm$ 0.05} & 74.70 \scalebox{0.7}{$\pm$ 0.07} \\
  & MGAT   & 53.74 \scalebox{0.7}{$\pm$ 0.48} & 46.93 \scalebox{0.7}{$\pm$ 0.85} & 81.40 \scalebox{0.7}{$\pm$ 0.13} & 72.27 \scalebox{0.7}{$\pm$ 0.31} & 80.19 \scalebox{0.7}{$\pm$ 0.25} & 77.30 \scalebox{0.7}{$\pm$ 0.25} & 78.11 \scalebox{0.7}{$\pm$ 0.08} & 74.15 \scalebox{0.7}{$\pm$ 0.08} \\
  & MIG-GT & 49.89 \scalebox{0.7}{$\pm$ 0.32} & 35.66 \scalebox{0.7}{$\pm$ 1.42} & 80.04 \scalebox{0.7}{$\pm$ 0.36} & 67.65 \scalebox{0.7}{$\pm$ 1.69} & 76.31 \scalebox{0.7}{$\pm$ 0.75} & 72.37 \scalebox{0.7}{$\pm$ 0.66} & 82.91 \scalebox{0.7}{$\pm$ 0.27} & 78.68 \scalebox{0.7}{$\pm$ 0.25} \\
  & NTSFormer & 54.07 \scalebox{0.7}{$\pm$ 0.51} & 46.91 \scalebox{0.7}{$\pm$ 0.36} & \second{83.47 \scalebox{0.7}{$\pm$ 0.18}} & \second{74.85 \scalebox{0.7}{$\pm$ 0.59}} & 79.91 \scalebox{0.7}{$\pm$ 0.20} & 77.84 \scalebox{0.7}{$\pm$ 0.18} & 86.13 \scalebox{0.7}{$\pm$ 0.03} & 82.81 \scalebox{0.7}{$\pm$ 0.09} \\ \cmidrule{1-10}

\multirow{2}{*}{\textbf{SUPRA}} 
  & Base & \second{55.92 \scalebox{0.7}{$\pm$ 0.15}} & \third{48.92 \scalebox{0.7}{$\pm$ 1.24}} & \third{83.38 \scalebox{0.7}{$\pm$ 0.26}} & 73.74 \scalebox{0.7}{$\pm$ 1.24} & \third{80.39 \scalebox{0.7}{$\pm$ 0.10}} & \third{78.13 \scalebox{0.7}{$\pm$ 0.13}} & \second{87.22 \scalebox{0.7}{$\pm$ 0.09}} & \first{84.51 \scalebox{0.7}{$\pm$ 0.13}} \\
  & + Aux   & \third{55.85 \scalebox{0.7}{$\pm$ 0.04}} & \first{49.81 \scalebox{0.7}{$\pm$ 1.30}} & \first{84.37 \scalebox{0.7}{$\pm$ 0.25}} & \first{76.71 \scalebox{0.7}{$\pm$ 0.11}} & \first{80.80 \scalebox{0.7}{$\pm$ 0.12}} & \first{79.16 \scalebox{0.7}{$\pm$ 0.12}} & \first{87.28 \scalebox{0.7}{$\pm$ 0.11}} & \second{84.42 \scalebox{0.7}{$\pm$ 0.07}} \\ \midrule \midrule

\rowcolor{gray!10}
\multicolumn{10}{c}{\textit{Feature Group: High-Confidence Priors (Llama-3.2)}} \\ \midrule

\multirow{3}{*}{\textbf{MLP}} 
  & Text   & 51.77 \scalebox{0.7}{$\pm$ 0.37} & 45.14 \scalebox{0.7}{$\pm$ 0.41} & 86.62 \scalebox{0.7}{$\pm$ 0.08} & \third{80.31 \scalebox{0.7}{$\pm$ 0.50}} & 81.06 \scalebox{0.7}{$\pm$ 0.18} & 79.77 \scalebox{0.7}{$\pm$ 0.07} & 58.49 \scalebox{0.7}{$\pm$ 0.05} & 56.54 \scalebox{0.7}{$\pm$ 0.18} \\
  & Visual & 53.72 \scalebox{0.7}{$\pm$ 0.33} & 45.29 \scalebox{0.7}{$\pm$ 0.70} & 73.40 \scalebox{0.7}{$\pm$ 0.27} & 66.59 \scalebox{0.7}{$\pm$ 0.21} & 73.48 \scalebox{0.7}{$\pm$ 0.21} & 72.00 \scalebox{0.7}{$\pm$ 0.35} & 78.74 \scalebox{0.7}{$\pm$ 0.06} & 74.16 \scalebox{0.7}{$\pm$ 0.11} \\
  & EF-MLP & \third{57.51 \scalebox{0.7}{$\pm$ 0.20}} & \third{51.55 \scalebox{0.7}{$\pm$ 0.67}} & 86.51 \scalebox{0.7}{$\pm$ 0.01} & 79.69 \scalebox{0.7}{$\pm$ 0.24} & 81.61 \scalebox{0.7}{$\pm$ 0.19} & 80.36 \scalebox{0.7}{$\pm$ 0.43} & 84.80 \scalebox{0.7}{$\pm$ 0.09} & 81.39 \scalebox{0.7}{$\pm$ 0.05} \\ \cmidrule{1-10}

\multirow{4}{*}{\textbf{GNNs}} 
  & GCN    & 55.47 \scalebox{0.7}{$\pm$ 0.49} & 50.09 \scalebox{0.7}{$\pm$ 0.19} & 84.51 \scalebox{0.7}{$\pm$ 0.19} & 75.63 \scalebox{0.7}{$\pm$ 0.34} & 81.56 \scalebox{0.7}{$\pm$ 0.30} & 79.44 \scalebox{0.7}{$\pm$ 0.57} & 79.32 \scalebox{0.7}{$\pm$ 0.01} & 75.62 \scalebox{0.7}{$\pm$ 0.04} \\
  & GAT    & 54.29 \scalebox{0.7}{$\pm$ 0.36} & 48.02 \scalebox{0.7}{$\pm$ 0.37} & 83.90 \scalebox{0.7}{$\pm$ 0.17} & 75.42 \scalebox{0.7}{$\pm$ 0.34} & 81.16 \scalebox{0.7}{$\pm$ 0.27} & 79.01 \scalebox{0.7}{$\pm$ 0.16} & 78.63 \scalebox{0.7}{$\pm$ 0.09} & 74.78 \scalebox{0.7}{$\pm$ 0.20} \\
  & SAGE   & 57.06 \scalebox{0.7}{$\pm$ 0.10} & 51.26 \scalebox{0.7}{$\pm$ 0.36} & 85.96 \scalebox{0.7}{$\pm$ 0.11} & 78.43 \scalebox{0.7}{$\pm$ 0.89} & 81.93 \scalebox{0.7}{$\pm$ 0.13} & 80.28 \scalebox{0.7}{$\pm$ 0.28} & 87.22 \scalebox{0.7}{$\pm$ 0.20} & \third{84.11 \scalebox{0.7}{$\pm$ 0.33}} \\
  & JKNet  & 55.75 \scalebox{0.7}{$\pm$ 0.12} & 50.12 \scalebox{0.7}{$\pm$ 0.87} & 84.07 \scalebox{0.7}{$\pm$ 0.28} & 74.47 \scalebox{0.7}{$\pm$ 0.33} & 81.33 \scalebox{0.7}{$\pm$ 0.02} & 79.08 \scalebox{0.7}{$\pm$ 0.18} & 79.45 \scalebox{0.7}{$\pm$ 0.12} & 75.55 \scalebox{0.7}{$\pm$ 0.10} \\ \cmidrule{1-10}

\multirow{4}{*}{\textbf{MAGs}} 
  & MMGCN  & 55.19 \scalebox{0.7}{$\pm$ 0.31} & 50.25 \scalebox{0.7}{$\pm$ 1.19} & 84.56 \scalebox{0.7}{$\pm$ 0.05} & 75.97 \scalebox{0.7}{$\pm$ 0.24} & 81.18 \scalebox{0.7}{$\pm$ 0.33} & 79.56 \scalebox{0.7}{$\pm$ 0.07} & 79.25 \scalebox{0.7}{$\pm$ 0.06} & 75.63 \scalebox{0.7}{$\pm$ 0.06} \\
  & MGAT   & 54.44 \scalebox{0.7}{$\pm$ 0.73} & 48.84 \scalebox{0.7}{$\pm$ 0.97} & 84.28 \scalebox{0.7}{$\pm$ 0.10} & 75.65 \scalebox{0.7}{$\pm$ 0.53} & 81.31 \scalebox{0.7}{$\pm$ 0.20} & 79.24 \scalebox{0.7}{$\pm$ 0.17} & 77.90 \scalebox{0.7}{$\pm$ 0.18} & 74.06 \scalebox{0.7}{$\pm$ 0.21} \\
  & MIG-GT & 55.93 \scalebox{0.7}{$\pm$ 1.68} & 46.13 \scalebox{0.7}{$\pm$ 1.78} & 85.82 \scalebox{0.7}{$\pm$ 0.55} & 76.67 \scalebox{0.7}{$\pm$ 1.82} & 81.36 \scalebox{0.7}{$\pm$ 0.50} & 78.96 \scalebox{0.7}{$\pm$ 0.28} & 84.69 \scalebox{0.7}{$\pm$ 0.29} & 80.38 \scalebox{0.7}{$\pm$ 0.18} \\
  & NTSFormer & 56.14 \scalebox{0.7}{$\pm$ 0.31} & 50.93 \scalebox{0.7}{$\pm$ 0.38} & \third{87.03 \scalebox{0.7}{$\pm$ 0.20}} & 80.13 \scalebox{0.7}{$\pm$ 0.25} & \third{82.60 \scalebox{0.7}{$\pm$ 0.21}} & \third{81.09 \scalebox{0.7}{$\pm$ 0.06}} & \third{87.28 \scalebox{0.7}{$\pm$ 0.11}} & 84.05 \scalebox{0.7}{$\pm$ 0.06} \\ \cmidrule{1-10}

\multirow{2}{*}{\textbf{SUPRA}} 
  & Base & \first{59.03 \scalebox{0.7}{$\pm$ 0.64}} & \second{54.15 \scalebox{0.7}{$\pm$ 0.50}} & \second{87.45 \scalebox{0.7}{$\pm$ 0.11}} & \second{80.83 \scalebox{0.7}{$\pm$ 0.40}} & \second{83.22 \scalebox{0.7}{$\pm$ 0.18}} & \second{81.73 \scalebox{0.7}{$\pm$ 0.09}} & \second{88.74 \scalebox{0.7}{$\pm$ 0.05}} & \first{86.11 \scalebox{0.7}{$\pm$ 0.17}} \\
  & + Aux & \second{58.90 \scalebox{0.7}{$\pm$ 0.32}} & \first{55.06 \scalebox{0.7}{$\pm$ 0.62}} & \first{87.57 \scalebox{0.7}{$\pm$ 0.14}} & \first{81.09 \scalebox{0.7}{$\pm$ 0.15}} & \first{83.33 \scalebox{0.7}{$\pm$ 0.30}} & \first{82.10 \scalebox{0.7}{$\pm$ 0.09}} & \first{88.75 \scalebox{0.7}{$\pm$ 0.05}} & \second{85.99 \scalebox{0.7}{$\pm$ 0.13}} \\ \bottomrule
\end{tabular}
}
\end{table*}

\textbf{Datasets \& Input Features.} We evaluate on four multimodal graph benchmarks from MAGB~\cite{yan2025graph} (Movies, Grocery, Toys, Reddit-M) spanning e-commerce and social media. Crucially, to empirically validate our theoretical SNR analysis (\S\ref{sec:theory}), we evaluate across two distinct feature quality regimes: (1) \textit{Standard features} from CLIP~\cite{CLIP} and RoBERTa~\cite{Roberta}, and (2) \textit{High-Confidence features} from Llama-3.2-11B-Vision~\cite{llamav2}.
Dataset statistics are provided in the Appendix (Table~\ref{tab:datasets}).

\textbf{Baselines \& Evaluation Protocol.} We report classification Accuracy and Macro-F1 score against three categories of methods: (1) topology-agnostic MLPs, (2) standard GNNs (GCN~\cite{GCN}, GraphSAGE~\cite{SAGE}, GAT~\cite{GAT}, JKNet~\cite{JKNet}), and (3) multimodal GNN architectures (MMGCN~\cite{mmgcn}, MGAT~\cite{mgat}, MIG-GT~\cite{MIGGT}, NTSFormer~\cite{ntsformer}). 
Due to space constraints, detailed experimental configurations including hyperparameter search spaces and additional results are deferred to Appendix~\ref{app:exp_details_results}.

\subsection{Main Results}
\label{sec:performance}

Table~\ref{tab:main_perf} reports the overall performance. Our analysis yields four critical insights:

\ding{172} \textbf{\model achieves universal state-of-the-art performance.}
\model demonstrates comprehensive dominance across diverse domains. Under High-Confidence Priors, \model with auxiliary supervision achieves top-tier results across all datasets (e.g., 88.75 on Reddit-M, 87.57 on Grocery), confirming the superiority of our decoupled paradigm.

\ding{173} \textbf{Performance Inversion validates SNR degradation (Theorem~\ref{thm:snr}).}
Under High-Confidence Priors, EF-MLP (86.51) surpasses SAGE (85.96) on Grocery, exhibiting 
clear Performance Inversion. This directly confirms Theorem~\ref{thm:snr}: when 
features are high-quality, mandatory aggregation introduces more noise than collaborative benefit.

\ding{174} \textbf{Ego-feature preservation outweighs complex cross-modal fusion.}
Canonical GNNs that retain ego-features (e.g., SAGE via root-concatenation) outperform 
sophisticated MAGs. On Standard Reddit-M, SAGE (86.44) and NTSFormer (86.13) substantially 
eclipse MMGCN (78.46). This exposes a fundamental flaw: GCN-style symmetric aggregation 
inherently mixes ego-features with neighborhood noise. \model resolves this by taking 
ego-feature preservation to the extreme via complete topological decoupling in its 
\textit{Unique Streams}.

\begin{figure*}[t]
  \centering
  \includegraphics[width=1.0\linewidth]{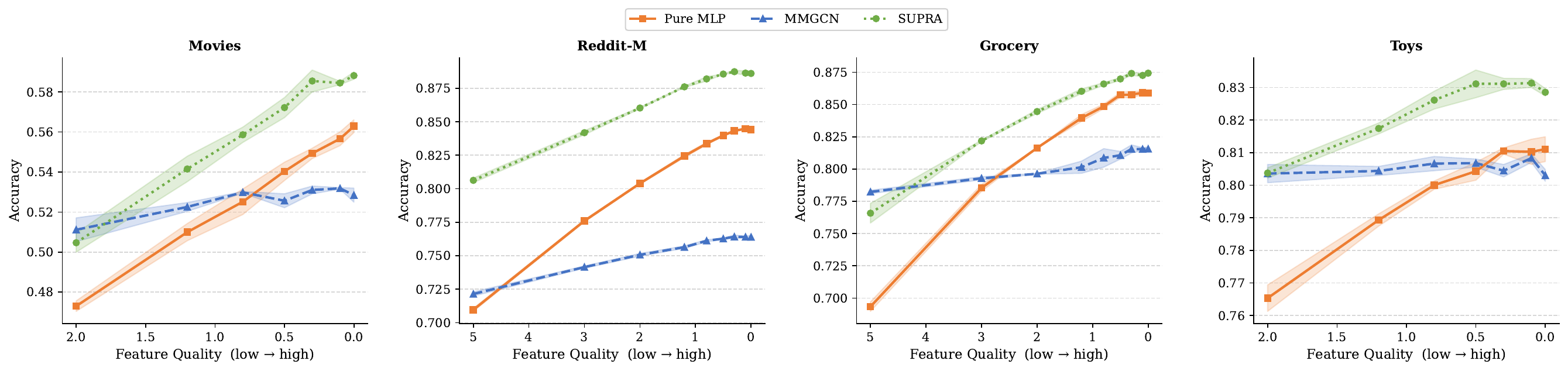}
  \vspace{-0.6cm}
  \caption{Performance dynamics under controlled feature quality regimes. The crossover between Pure MLP and MMGCN validates the SNR degradation threshold ($\tau(\beta)$).}
  \label{fig:snr_crossover}
  \vspace{-0.4cm}
\end{figure*}

\begin{figure*}[t]
  \centering
  \includegraphics[width=1.0\linewidth]{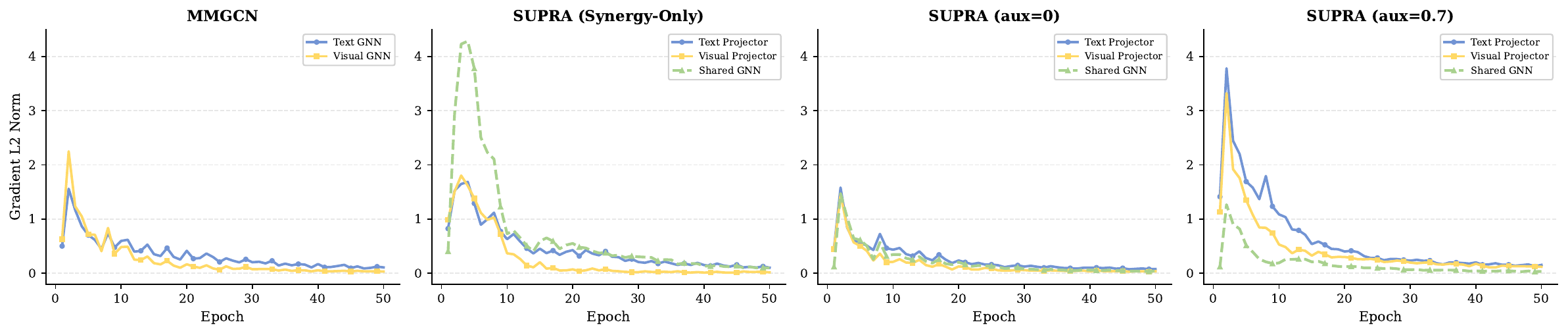}
  \caption{Gradient L2 norm evolution during training on the Grocery dataset across four architectural variants. \model (aux=0.7) uniquely overcomes both architectural and optimization bottlenecks, maintaining robust, sustained gradients for the modality-specific projectors while successfully preventing the shared GNN from dominating the optimization dynamics.}
  \label{fig:grad_starvation}
\end{figure*}

\subsection{Validating Theorem 1: The SNR Crossover Phenomenon}
\label{sec:exp_noise}

To empirically validate the representational pathology defined in Theorem~\ref{thm:snr}, we systematically modulate the intrinsic feature quality ($\sigma_\epsilon^2$) by injecting scaled Gaussian noise into the initial foundation model embeddings.
We track the performance of a topology-agnostic model (MLP), a mandatory-aggregation model (MMGCN), and our decoupled architecture (\model Base, isolating the SNR preservation mechanism without auxiliary loss).

\textbf{Results.} As feature quality increases (moving left to right in Figure~\ref{fig:snr_crossover}), the observed shift perfectly aligns with our mathematical bounds. \ding{172} \textbf{Performance crossover validates the $\tau(\beta)$ threshold:} In the low-quality regime, structural aggregation acts as a crucial smoothing mechanism, allowing MMGCN to significantly outperform the MLP baseline. However, as feature quality improves, a sharp crossover occurs, whereby the MLP systematically eclipses MMGCN.  This confirms the critical threshold $\tau(\beta)$: in the \textit{High-Confidence regime} ($\sigma_\epsilon^2 < \tau(\beta)$), mandatory aggregation introduces topological noise that overwhelms the high-fidelity semantic signal. 
\ding{173} \textbf{\model consistently recovers the global upper bound:} While MMGCN exhibits a slight advantage at extreme noise levels, \model demonstrates robust adaptation across the meaningful quality spectrum. 
Past the crossover point, its Unique Specificity streams shield high-fidelity semantics from topological dilution.
Unlike MMGCN which plateaus, \model tracks the steep upward trajectory of MLP, consistently achieving the global upper bound across all evaluated datasets.

\subsection{Verifying Optimization Dynamics: Gradient Starvation}
\label{sec:exp_grad}
To verify Proposition~\ref{prop:starvation}, we track the gradient $L_2$ norms flowing into the modality projectors on Grocery (Figure~\ref{fig:grad_starvation}).
Four architectures disentangle the concurrent theoretical bottlenecks: (1) {MMGCN} (coupled baseline); (2) {\model (Synergy-Only)}, which isolates the \textit{Architectural Bottleneck} by forcing all signals through the GNN (expected attenuation $\eta = \mathcal{O}(\alpha^L)$); 
(3) {\model (aux=0)}, isolating the {Optimization Bottleneck} by utilizing a {topology-agnostic} bypass ($\eta = 1$) but remaining under a shared task loss ($r_i(t) \to 0$); and (4) {\model (aux=0.7)}, our full decoupled paradigm.

\textbf{Results.} The observed degradation chain aligns closely with our theoretical predictions. 
\ding{172} \textbf{Architectural bottleneck induces rapid decay:} Forcing representations exclusively through the shared GNN in MMGCN and \model (Synergy-Only) causes the gradient of the weaker visual modality to diminish rapidly, visually confirming the $\eta$ attenuation effect. \ding{173} \textbf{Shared objective triggers premature starvation:} Introducing a topology-agnostic bypass (\model aux=0) provides an initial gradient spike; however, the shared task loss ($\mathcal{L}_{\text{task}}$) allows the dominant modality to rapidly minimize the joint residual, subsequently causing premature starvation. 
\ding{174} \textbf{Gradient isolation sustains modality learning:} Applying explicit auxiliary rewiring (\model aux=0.7) fundamentally alters the optimization dynamics. Both projectors maintain substantial and sustained gradients. Interestingly, the shared GNN gradient remains substantially {lower} than the projectors, indicating that \model prioritizes intrinsic semantics and utilizes topology strictly as a collaborative supplement.

\begin{figure*}[t!] 
  \centering
  \begin{minipage}{0.5\linewidth}
    \centering
    \includegraphics[width=\linewidth]{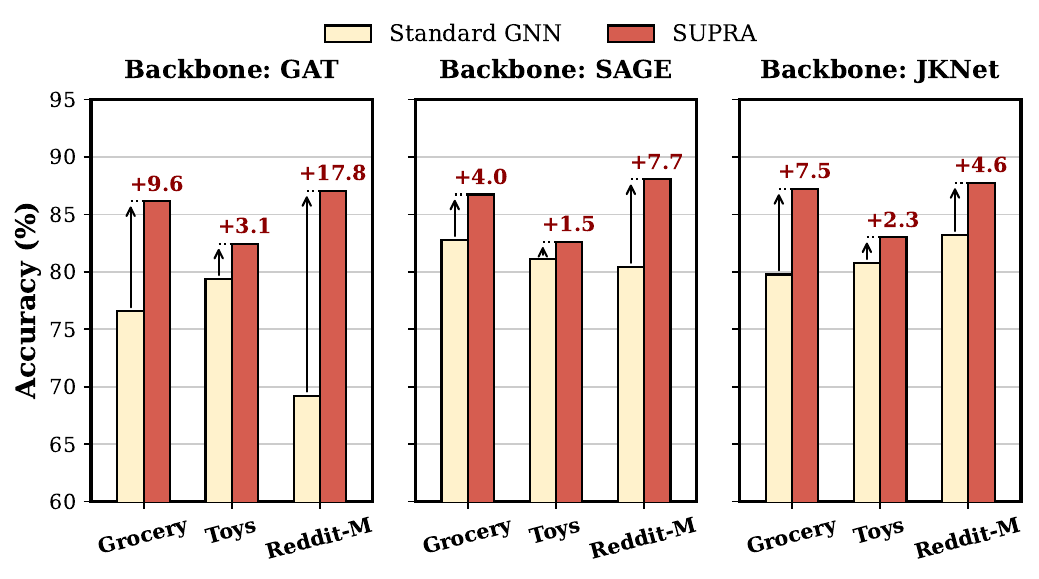} 
    \centerline{(a) Accuracy}
  \end{minipage}\hfill
  \begin{minipage}{0.5\linewidth}
    \centering
    \includegraphics[width=\linewidth]{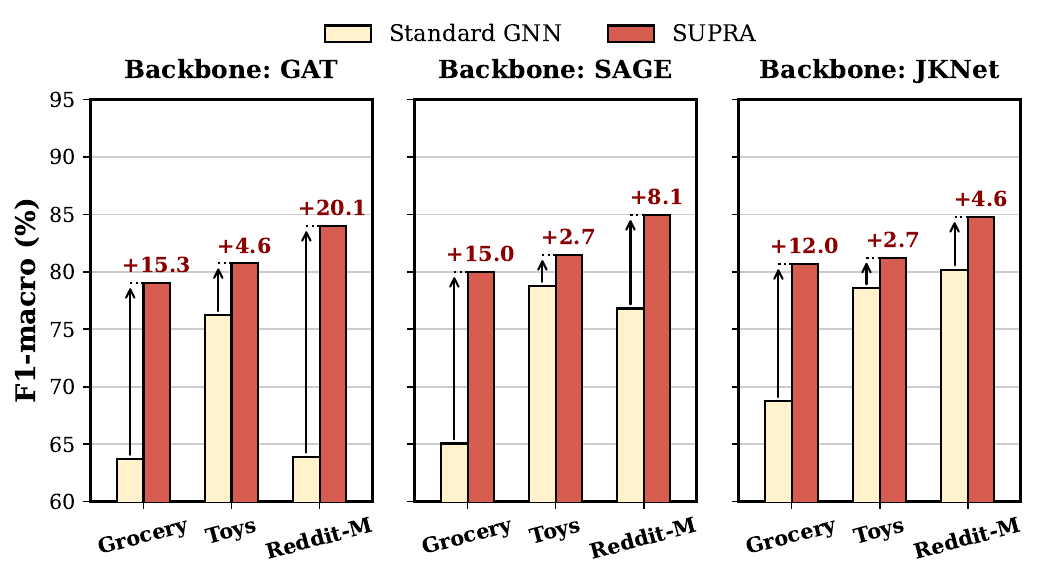} 
    \centerline{(b) Macro-F1}
  \end{minipage}
  
  \caption{Generalizability across diverse message-passing backbones.}
  \label{fig:backbone}
  \vspace{-0.4cm}
\end{figure*}

\begin{figure}[t!]
  \centering
  \includegraphics[width=\linewidth]{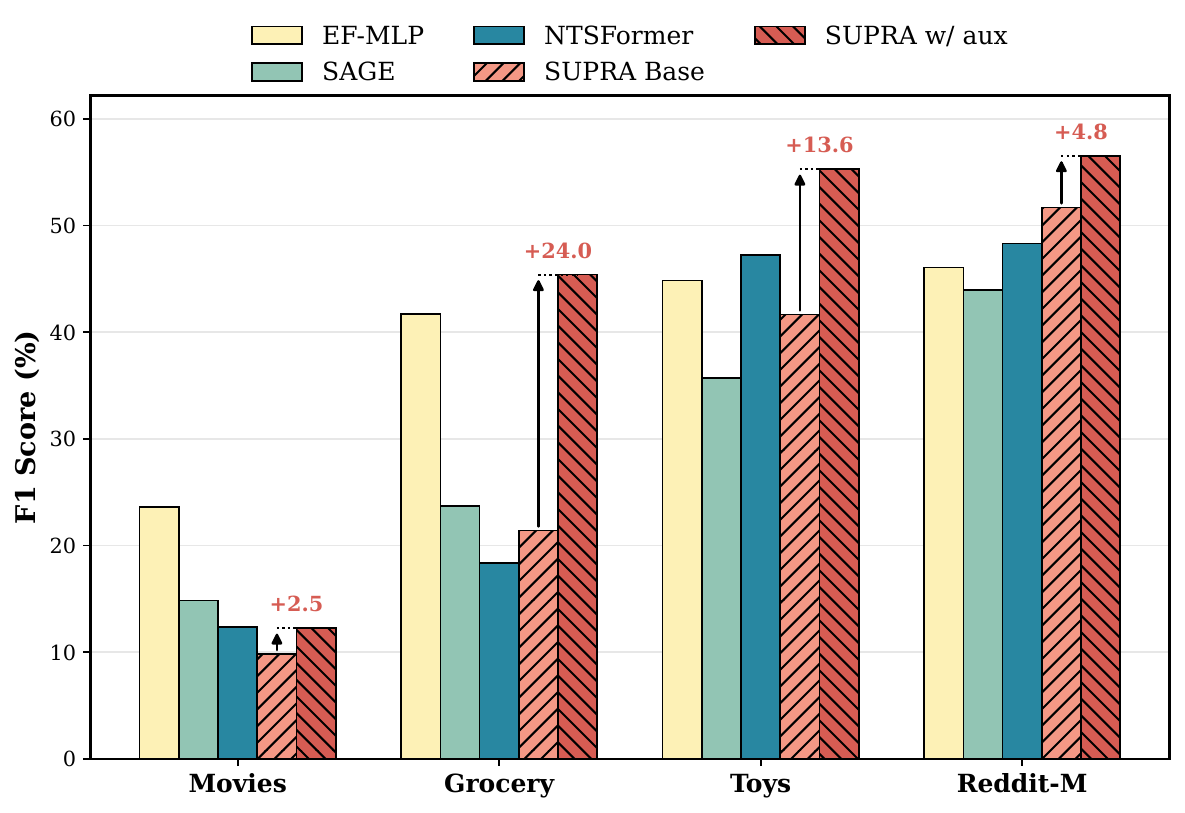}
  \caption{Robustness under modality degradation. The dominant modality is corrupted at test time to enforce reliance on weaker modalities.}
  \label{fig:degrade}
\end{figure}

\subsection{Diagnosing Gradient Starvation via Modality Corruption}
\label{sec:exp_degrade}

Having verified the internal optimization dynamics during training (\S\ref{sec:exp_grad}), we further design a diagnostic probe to evaluate its external consequences at inference time. By aggressively corrupting the dominant modality during testing, we force reliance on the weaker modality, revealing whether it was genuinely optimized or merely overshadowed.

\textbf{Results.} As shown in Figure~\ref{fig:degrade}, conventional models (such as SAGE and NTSFormer) experience severe performance degradation. Crucially, the performance of \model (Base) also collapses (by approximately 21\% on the Grocery dataset). This outcome demonstrates that \textit{Topological Decoupling} alone is insufficient: under a shared task loss ($\mathcal{L}_{\text{task}}$), the dominant modality rapidly minimizes the joint optimization objective, drives the residual to zero ($r(t) \to 0$), and prematurely starves the weaker projector.
Conversely, \model with auxiliary supervision reverses this collapse. Applying auxiliary loss ($\mathcal{L}_{\text{aux}}$) directly to the Unique Streams provides topology-agnostic optimization signals for the weaker modality, yielding substantial gains over the Base architecture on Grocery (+24.0\%) and Toys (+13.6\%). This confirms the necessity of our optimization design.

\subsection{Generality Across Diverse GNN Backbones}
\label{sec:exp_backbone}

To verify that the superiority of \model is not merely an artifact of mitigating specific limitations of GCN, we replace the graph encoder in the Synergy Stream with more advanced backbones, including GAT~\cite{GAT}, GraphSAGE~\cite{SAGE}, and JKNet~\cite{JKNet}.

\textbf{Results.}Figure~\ref{fig:backbone} demonstrates the backbone-independent effectiveness of \model: \ding{172} \textbf{Narrowing the architectural performance gap:} Under the standard paradigm, different GNN operators exhibit significant variance due to their varying sensitivity to topology. In contrast, \model elevates all evaluated backbones to a consistently high-performance regime, demonstrating that macroscopic decoupling effectively neutralizes the inherent vulnerabilities of coupled architectures. \ding{173} \textbf{Unhindered learning via explicit gradient isolation:} We observe pronounced improvements in both Accuracy and Macro-F1, with F1 gains being particularly notable (e.g., +20.1\% for GAT on Reddit-M). This corroborates our \textit{Propagation Dilemma} theory (\S\ref{sec:theory_dilemma}): rather than forcing distinct modalities through an entangled topological bottleneck, \model shields intrinsic modality signals from the shared collaborative channel. This prevents the joint gradient flow from suppressing weaker modalities, allowing each to faithfully optimize its features without interference.

\begin{figure*}[t]
    \centering
    \includegraphics[width=1.0\textwidth]{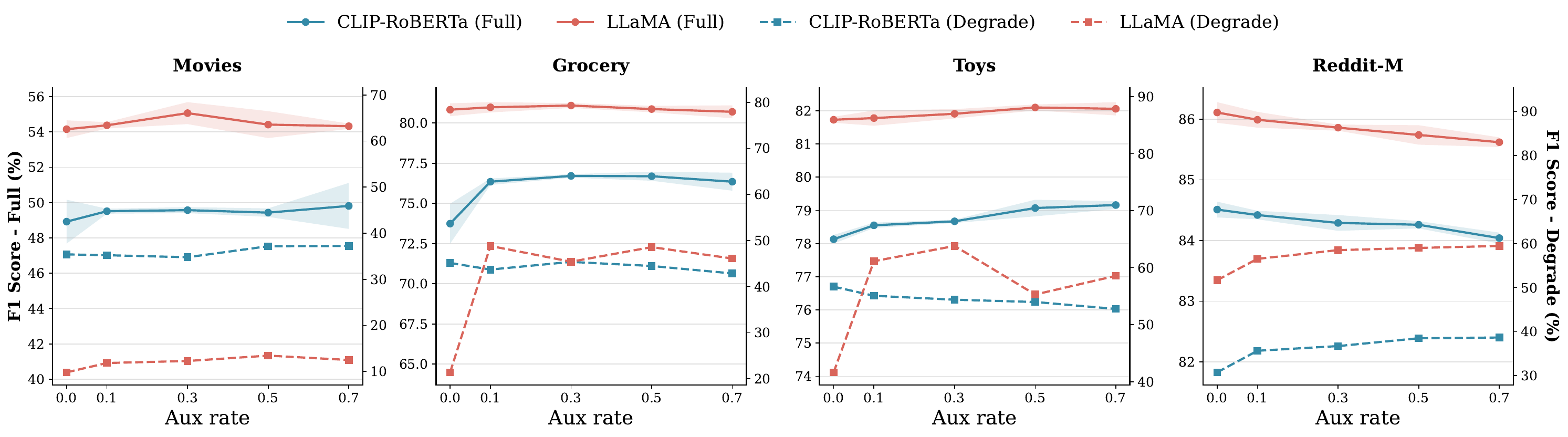}
    \caption{Performance sensitivity to the auxiliary supervision ratio $\lambda_{\text{aux}}$ across different datasets and feature priors. Solid lines denote the full modality setting, while dashed lines represent the modality degradation setting.}
    \label{fig:f1_aux_rate}
\end{figure*}

\begin{table*}[t!]
\centering
\caption{Comprehensive evaluation of model performance (F1-score) on full and degraded modalities. $F$: Full modality performance; $D$: Degraded modality performance; $H$: Harmonic Mean of $F$ and $D$. Best results are highlighted in bold.}
\label{tab:f1-robustness}
\resizebox{\textwidth}{!}{
\begin{tabular}{l ccc ccc ccc ccc}
\toprule
\multirow{2}{*}{\textbf{Method}} & \multicolumn{3}{c}{\textbf{Movies}} & \multicolumn{3}{c}{\textbf{Grocery}} & \multicolumn{3}{c}{\textbf{Toys}} & \multicolumn{3}{c}{\textbf{Reddit-M}} \\
\cmidrule(lr){2-4} \cmidrule(lr){5-7} \cmidrule(lr){8-10} \cmidrule(lr){11-13}
 & $F$ & $D$ & $H$ & $F$ & $D$ & $H$ & $F$ & $D$ & $H$ & $F$ & $D$ & $H$ \\
 \midrule
SUPRA      & 53.00 & 8.31 & 14.37 & \textbf{81.23} & 12.43 & 21.56 & 80.91 & 37.72 & 51.45 & 84.36 & 49.86 & 62.68 \\
+Aux & \textbf{53.47} & \textbf{11.02} & \textbf{18.27} & 80.77 & \textbf{40.26} & \textbf{53.74} & \textbf{81.47} & \textbf{56.04} & \textbf{66.40} & 83.49 & \textbf{60.13} & \textbf{69.91} \\
+OGM-GE     & 52.22 & 7.83 & 13.62 & 81.13 & 9.98 & 17.77 & 81.32 & 35.60 & 49.52 & \textbf{84.48} & 50.34 & 63.09 \\
\bottomrule
\end{tabular}
}
\end{table*}

\subsection{Sensitivity Analysis of Auxiliary Supervision Ratio $\lambda_{\text{aux}}$}
\label{sec:aux_ratio_analysis}

To further analyze the optimization dynamics of SUPRA, we investigate the impact of the auxiliary supervision weight $\lambda_{\text{aux}}$ on the overall performance of SUPRA, as illustrated in Figure \ref{fig:f1_aux_rate}. We conduct experiments across four datasets (Movies, Grocery, Toys, and Reddit-M) using two types of feature priors: CLIP-RoBERTa and LLaMA-Vision. The performance is evaluated under both the full modality setting and the modality degradation setting. Our empirical analysis yields the following key observations:

\ding{172} \textbf{Superiority of High-Confidence Priors.} Under the full modality setting, models utilizing LLaMA-Vision features outperform those relying on CLIP-RoBERTa across all datasets. This confirms that LLaMA-Vision serves as a high-confidence prior, providing a stronger semantic foundation for classification.

\ding{173} \textbf{Effectiveness of Auxiliary Supervision against Degradation.} Incorporating auxiliary supervision ($\lambda_{\text{aux}} > 0$) generally yields superior performance compared to the base architecture ($\lambda_{\text{aux}} = 0$). Crucially, these gains are most pronounced under the modality degradation setting (dashed lines). This confirms that explicitly supervising the Unique Streams provides independent gradient pathways, preventing premature gradient starvation of the weaker modality and ensuring robustness when the dominant prior is corrupted.

\ding{174} \textbf{Dataset-Specific Optimal Ratios.} The optimal value for $\lambda_{\text{aux}}$ varies depending on the dataset. For instance, the F1 scores typically peak around $\lambda_{\text{aux}} = 0.3$ or $0.5$ for datasets like Grocery and Toys. This variance highlights the necessity of tuning the auxiliary ratio to accommodate the distinct intrinsic multimodal dynamics of different graphs.

\ding{175} \textbf{Modality Interference on Reddit-M.} Unlike other datasets, increasing the auxiliary ratio on Reddit-M gradually degrades performance under the full modality setting. This indicates that its visual features possess significantly higher signal quality than the textual modality. Consequently, forcing the model to over-optimize the much weaker text features introduces optimization conflicts. We provide direct visual evidence for this severe modality disparity in the following section (Figure~\ref{fig:tsne}).

\subsection{Comparison with General Multimodal Gradient Balancing}
\label{sec:ogm_comparison}

To address modality imbalance, common practices in multimodal learning often involve dynamic gradient modulation, such as OGM-GE~\cite{OGMGE}. OGM-GE adaptively controls the optimization of each modality by monitoring their contribution discrepancy to the learning objective. To investigate whether these established solutions are directly applicable to Multimodal Attributed Graphs (MAGs), we compare our auxiliary supervision approach with OGM-GE. The comprehensive results, reporting Full modality performance ($F$), Degraded modality performance ($D$), and their Harmonic Mean ($H$), are summarized in Table~\ref{tab:f1-robustness}.
Our empirical comparison leads to the following observations regarding the application of general gradient balancing to graph data:

\textbf{Incompatibility with Graph Topology.} While OGM-GE is designed to balance modalities at the fusion stage, it shows limited efficacy in the graph domain, particularly under the modality degradation setting ($D$). For example, on the Grocery dataset, OGM-GE fails to improve upon the base performance, yielding only 9.98\% in $D$, whereas our approach reaches 40.26\%. This suggests that simple gradient rescaling at the final output layer may be insufficient when features are already deeply entangled with graph topology during the message-passing phase.

\textbf{Methodological Discussion.} It is worth noting that while auxiliary supervision is a straightforward implementation compared to the on-the-fly modulation of OGM-GE, it effectively bypasses the ``gradient starvation" caused by topological noise. However, we acknowledge that \model relies on a static hyperparameter $\lambda_{\text{aux}}$, whereas OGM-GE attempts a more dynamic balancing act. We consider the development of more sophisticated, graph-aware dynamic balancing mechanisms that account for structural propagation to be a promising direction for future research.

\subsection{Efficiency Analysis}
\label{sec:exp_efficiency}

We evaluate computational overhead on Reddit-M to assess the practical viability of \model.
We report Peak Reserved Memory ($\mathcal{M}_{\text{res}}$), measured via \textit{nvidia-smi}.
\ding{172} \textbf{Lower memory footprint.} \model achieves 3.5$\times$ lower memory than 
Graph Transformers (e.g., NTSFormer, MIG-GT) and 1.7$\times$ lower than basic GCN by propagating 
projected representations rather than raw foundation model features through the GNN.
\ding{173} \textbf{Efficient training dynamics.} Although \model requires more epochs than 
MMGCN (226 vs. 120), it converges 2.6$\times$ faster in wall-clock time than SAGE and 
4.4$\times$ faster than NTSFormer.
This performance gap highlights a fundamental distinction: the rapid convergence of MMGCN stems from early saturation, where the aggregation of topological noise prematurely stalls the optimization of modality-specific patterns. In contrast, \model sustains the extraction of discriminative features throughout the training process, achieving superior final performance with high temporal efficiency.
These results confirm that \model is a scalable solution that provides memory and training efficiency superior to or comparable with basic GNNs while preserving the representational quality necessitated by LFM priors.

\begin{table}[t]
  \centering
  \small
\caption{\textbf{Efficiency on Reddit-M.} $\mathcal{M}_{\text{res}}$: Peak Reserved GPU Memory (MB); $\mathcal{T}_{\text{ep}}$: Time (s/ep).}
\label{tab:efficiency}
\resizebox{\linewidth}{!}{
\begin{tabular}{lcccc}
\toprule
\textbf{Method} & \textbf{$\mathcal{M}_{\text{res}}$} & \textbf{$\mathcal{T}_{\text{ep}}$} & \textbf{$N_{\text{ep}}$} & \textbf{$\mathcal{T}_{\text{tot}}$} \\
\midrule
{\model}     & {7,956} & 0.035 & 226 & 12.7 \\
GCN                 & 13,199         & {0.032} & 185 & {9.3} \\
SAGE                & 10,416         & 0.036 & 532 & 32.8 \\
MMGCN               & 20,044         & 0.041 & 120 & 7.6 \\
NTSFormer           & 27,692         & 0.179 & 235 & 55.7 \\
MIG-GT              & 26,854         & 0.183 & 597 & 151.5 \\
\bottomrule
\end{tabular}
}
\vspace{-0.3cm}
\end{table}

\section{Related Work}
\label{sec:relatedwork}

\subsection{Multimodal Attributed Graph Learning}
Research in Multimodal Attributed Graph Learning (MAGL) has steadily progressed from naive early-fusion to specialized architectures. The first wave of MAGL models, such as \textbf{MMGCN}~\cite{mmgcn} and \textbf{MGAT}~\cite{mgat}, introduced modality-separated pathways to process distinct features over separate structures. More recently, Multimodal Graph Transformers (e.g., \textbf{MIG-GT}~\cite{MIGGT}, \textbf{NTSFormer}~\cite{ntsformer}) have employed heavily parameterized attention mechanisms to capture complex cross-modal interactions. While these models focus on fine-grained fusion, they implicitly rely on the assumption that structural aggregation is always beneficial, often leading to mandatory, dense topological mixing that risks diluting robust intrinsic semantics under high-confidence priors.
Beyond specialized MAGL models, several general-purpose Graph Neural Networks have explored mechanisms to mitigate over-smoothing and preserve the influence of intrinsic node attributes. \textbf{GraphSAGE}~\cite{SAGE} utilizes \textit{concatenation-based aggregation} to retain the node's current representation, while \textbf{JKNet}~\cite{JKNet} employs \textit{layer-wise skip connections} to selectively aggregate multi-hop information, effectively allowing the model to favor the original features.However, despite their feature-preserving intent, these architectures are still ill-equipped for the LFM era in MAGs for two primary reasons: (1) Structural Noise Interference: they typically operate on a coupled propagation-aggregation framework where even a single layer of message passing can contaminate high-confidence LFM priors with topological noise; and (2) Optimization Bottleneck: they lack independent supervision pathways, remaining susceptible to the \textit{gradient starvation} identified in Section~\ref{sec:theory_grad}, where dominant modalities suppress the learning of weaker branches regardless of skip connections. Consequently, existing paradigms still struggle to resolve the fundamental {aggregation dilemma}.

\subsection{Foundation Models for Graph Learning}
The integration of Large Foundation Models (LFMs) has emerged as a transformative trend in graph learning, generally categorized into two paradigms. The first, \textit{LFM-enhanced GNNs}~\cite{llm_semantic,Exploring,CS-TAG}, utilizes LFMs as robust feature extractors for downstream GNNs. However, directly feeding these high-quality features into dense message-passing mechanisms risks diluting intrinsic semantics with neighborhood noise. The second paradigm, \textit{Graph-enhanced LFMs}, positions the language model as the core reasoner, integrating structural information via textual prompting. While recent prompt-engineering efforts like DGP~\cite{DGP} attempt to mitigate information overload by filtering and summarizing neighborhood contexts, relying on extensive LLM inference for structural reasoning incurs prohibitive computational costs. Consequently, the community faces a critical dilemma: existing paradigms struggle with either semantic entanglement due to structural coupling, or excessive computational overhead from exhaustive graph prompting. This highlights an urgent need for lightweight architectures capable of effectively preserving the purity of strong LFM priors.

\section{Conclusion}\label{sec:conclusion}

In this work, we investigated the counter-intuitive phenomenon of \textit{performance inversion} in Multimodal Attributed Graph Learning (MAGL), where simple MLPs outperform sophisticated graph models when equipped with high-confidence LFM priors. We demonstrated that this anomaly stems from a fundamental {structural compromise} in existing paradigms: mandatory topological aggregation structurally degrades the Signal-to-Noise Ratio (SNR) of strong semantic signals and induces a dual-bottleneck optimization pathology termed {Gradient Starvation}. To resolve these issues, we propose {\model}, a {decoupled dual-pathway architecture} based on the principle of {selective propagation}. By isolating node-level intrinsic multimodal attributes into topology-agnostic specificity streams and capturing structural context via a lightweight synergy stream, \model successfully reconciles the priority of high-confidence semantic priors with the utility of topological consensus. Extensive evaluations confirm that \model counteracts both representational and optimization pathologies, achieving state-of-the-art performance with remarkably low computational overhead. 

\section*{Impact Statement} 
This work advances the sustainability and equity of Multimodal Attributed Graph Learning (MAGL). By drastically reducing memory and training costs, our framework democratizes access for resource-constrained institutions and minimizes the carbon footprint of AI training. Ethically, resolving ``Gradient Starvation'' prevents the marginalization of weaker but critical modalities, ensuring a balanced and robust integration of diverse data sources. Ultimately, this provides a reliable foundation for the responsible deployment of LFM-driven graph technologies in sensitive real-world domains.

\clearpage
\bibliographystyle{unsrt}
\bibliography{sample-base}

\appendices
\onecolumn

\section{Experimental Details and Additional Results}
\label{app:exp_details_results}

\subsection{Baseline Algorithms}
To provide a comprehensive evaluation, we compare SUPRA against 11 representative baselines categorized into three distinct paradigms:

\begin{itemize}
    \item \textbf{Topology-Agnostic Baselines (MLP):} These methods ignore graph topology entirely to measure the raw representation power of the node features. We evaluate unimodal MLPs (\textbf{Text} and \textbf{Visual}) and a multimodal version that concatenates the multimodal features before feeding them into a shared MLP.
    
    \item \textbf{Standard Graph Neural Networks (GNNs):} These methods represent traditional topological propagation applied to early-fused (concatenated) multimodal features. We benchmark against \textbf{GCN}, \textbf{GraphSAGE} (SAGE), \textbf{GAT}, and \textbf{JKNet} to evaluate standard homophily-based aggregation strategies on the unified feature space.
    
    \item \textbf{Multimodal Graph Learning Methods:} These are architectures explicitly designed to model Multimodal Attributed Graphs (MAGs). We compare against \textbf{MMGCN}, \textbf{MGAT}, \textbf{MIG-GT}, and the recently introduced \textbf{NTSFormer}, which serves as a strong Transformer-based baseline for multimodal graph representation.
    
\end{itemize}

\subsection{Implementation Details}
All methods are implemented using \textbf{PyTorch} and the \textbf{Deep Graph Library (DGL)}.
To ensure a fair comparison, we adhere to the following configurations:
\begin{itemize}
    \item \textbf{Hardware \& Compute Profiling:} All experiments are conducted on a workstation equipped with a single NVIDIA GeForce RTX 5090 GPU.
    \item \textbf{Data Splits:} For all evaluated datasets, we apply an identically fixed 60\%/20\%/20\% split for the training, validation, and testing sets across all baseline models to ensure a fair and completely reproducible comparison. The validation set is used exclusively for hyperparameter tuning and early stopping, while the final performance metrics are reported solely on the unseen test set.
    \item \textbf{Architecture:} We fix the hidden dimension for all neural backbones to $d=256$. To mitigate the over-smoothing issue typical of standard GNNs, the number of layers $L$ for GCN and GAT is restricted to $L \in \{1, 2\}$. Conversely, for architectures that are more robust to structural depth, such as GraphSAGE, JKNet and our proposed SUPRA, we expand the search space to $L \in \{2, 3, 4\}$.
    \item \textbf{Optimization:} We use the Adam optimizer with a weight decay of $1e^{-4}$. To improve generalization, label smoothing is applied with a factor of $0.1$ across all experiments. Dropout rate is fixed at $0.3$.
    \item \textbf{Reporting:} We report the mean accuracy and Macro-F1 score, along with their standard deviations, over 3 independent runs using different random seeds.
\end{itemize}

\subsection{Hyperparameter Configuration}
We perform a grid search to determine the optimal hyperparameters for all methods, selecting the best model based on validation performance with an early stopping patience of 40 to 50 epochs:
\begin{itemize}
    \item \textbf{General Settings:} The initial learning rate is tuned within $\{5e^{-4}, 1e^{-3}\}$ for standard GNN and MLP backbones.
    \item \textbf{Baseline Specifics:} For traditional GNNs and MLP-based methods, we tune the standard parameters as described above. For the advanced Multimodal Graph Learning architectures (i.e., MIG-GT and NTSFormer), the learning rate search space is expanded to include $5e^{-3}$. Furthermore, we directly adopt the method-specific hyperparameters (e.g., the number of attention heads, $k_t$/$k_v$ dimensions, and specific Transformer layers) as recommended in their respective original papers to ensure optimal and fair baseline performance.
    \item \textbf{SUPRA Specifics:} We evaluate two variants: \textbf{SUPRA (Base)} without auxiliary supervision ($\lambda_{aux} = 0.0$), and \textbf{SUPRA (+Aux)} with $\lambda_{aux}$ tuned in $\{0.1, 0.3, 0.5, 0.7\}$. Additionally, the modality-specific base projectors ($f_m$) are implemented as single-layer perceptrons (with ReLU activations) that map the input features to a unified projection dimension, matching the GNN hidden dimension ($d=256$), prior to concatenation.
\end{itemize}

\subsection{Visualization of Modality Disparity.} 
\label{sec:vis_disparity}

\begin{figure*}[t] 
    \centering
    \includegraphics[width=1\linewidth]{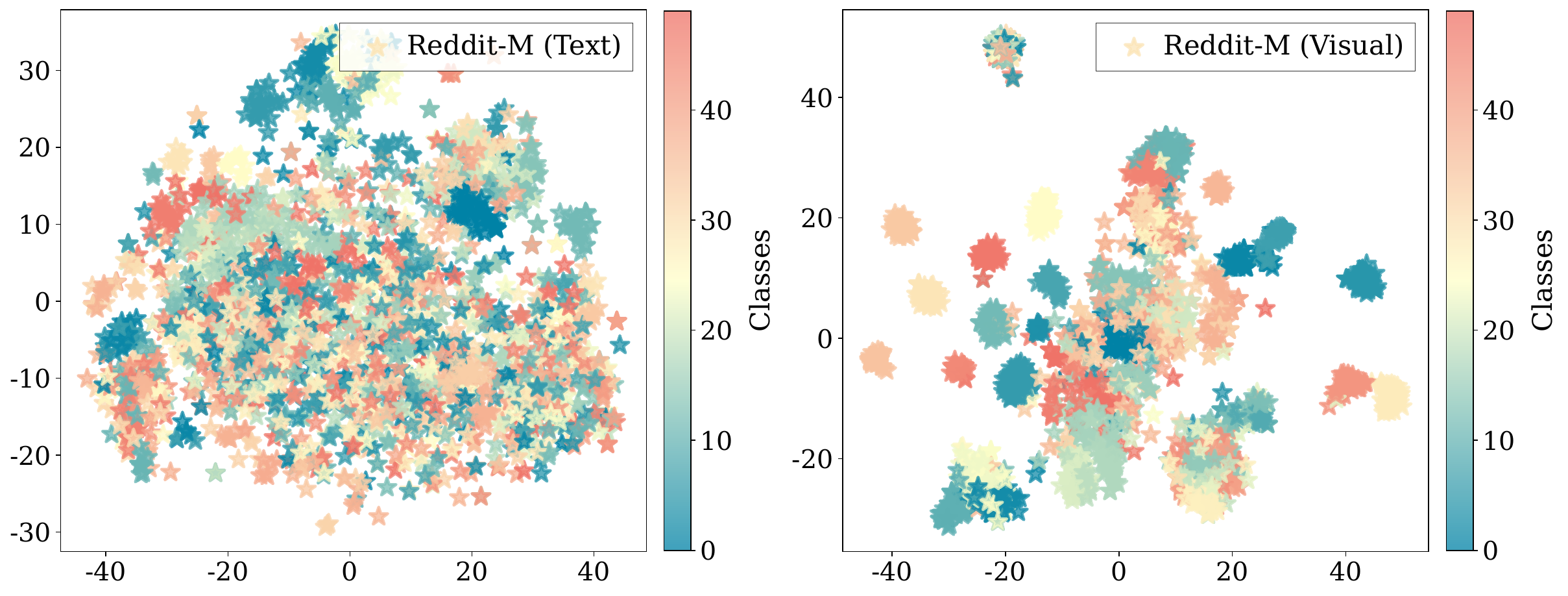} 
    \caption{T-SNE visualization of Textual and Visual features from Llama-Vision on Reddit-M.}
    \label{fig:tsne}
\end{figure*}

Figure~\ref{fig:tsne} visualizes the t-SNE embeddings of the features extracted by Llama-3.2-11B-Vision on the Reddit-M dataset. A sharp contrast is observed: the visual features (right) form compact and distinct clusters, indicating a high signal-to-noise ratio (SNR). Conversely, the textual features (left) exhibit severe overlap and lack clear class boundaries, which suggests a low SNR. This \textit{intrinsic learnability gap} provides an explanation for modality laziness, wherein coupled models naturally gravitate toward the visual manifold that is easier to learn~\cite{modalitylazy}. Furthermore, this visual evidence empirically validates the necessity of the proposed architecture of \model. By implementing \textit{Topological Decoupling} and explicit \textit{Gradient Isolation}, \model effectively prevents the noisy textual representations from compromising the clean visual signals, while ensuring that the weaker textual modality is genuinely optimized rather than prematurely suppressed.

\begin{table}[t]
    \centering
    \caption{Summary of Key Notations.}
    \label{tab:notations}
    \footnotesize
    \setlength{\tabcolsep}{5pt}
    \renewcommand{\arraystretch}{1.05}

    \begin{tabular}{p{0.28\linewidth} p{0.62\linewidth}}
        \toprule
        \textbf{Notation} & \textbf{Description} \\
        \midrule

        \multicolumn{2}{c}{\textit{Graph \& Features}} \\
        
        $\mathcal{G}=(\mathcal{V},\mathcal{E},\mathcal{X})$
        & Multimodal attributed graph \\

        $N$
        & Number of nodes \\

        $C$
        & Number of classes \\

        $\mathbf{A}\in\{0,1\}^{N\times N}$
        & Adjacency matrix \\

        $\mathcal{M}$
        & Set of modalities (e.g., $\{t,v\}$) \\

        $\mathbf{X}^{(m)}\in\mathbb{R}^{N\times d_m}$
        & Feature matrix for modality $m$ \\

        $\mathbf{x}_v^{(m)}$
        & Feature vector of node $v$ under modality $m$ \\

        $\mathcal{N}_v$
        & Neighbor set of node $v$ \\

        \midrule

        \multicolumn{2}{c}{\textit{Model Architecture \& Optimization}} \\

        $\mathbf{Z}\in\mathbb{R}^{N\times C}$
        & Learned node representations (logits) \\

        $\mathbf{H}^{(S)},\mathbf{H}^{(U_m)}$
        & Shared synergy and unique specificity embeddings \\

        $g_s(\cdot)$
        & Bottleneck MLP for dimensionality reduction \\

        $F_\theta(\cdot)$
        & Mapping function parameterized by $\theta$ \\

        $\mathcal{L}_{\text{task}}$
        & Shared task prediction loss \\

        \midrule

        \multicolumn{2}{c}{\textit{Theoretical Analysis}} \\

        $\alpha\in(0,1)$
        & Self-retention weight in mean aggregation \\

        $\beta\in[0,1]$
        & Semantic alignment coefficient \\

        $\tau(\beta)$
        & Critical SNR degradation threshold \\

        \bottomrule
    \end{tabular}
\end{table}

\begin{table}[t]
  \centering
  \setlength{\abovecaptionskip}{0.2cm}
  \small
  \captionsetup{skip=5pt} 
  \caption{Basic statistics of MAGs in the MAGB. The variable $L_{\rm text}$ represents the average number of words, while $R_{\rm img}$ denotes the resolution of the image (in pixels).}
    \begin{tabular}{lcccc}
    \toprule
    Dataset & Movies & Toys & Grocery & Reddit-M \\
    \midrule
    \#Nodes & 16,672 & 20,695 & 17,074 & 99,638 \\
    \#Edges & 218,390 & 126,886 & 171,340 & 1,167,188 \\
    \#Class & 20 & 18 & 20 & 50 \\
    $H_{\rm hom}$ & 47.56\% & 73.04\% & 70.07\%  & 67.16\% \\
    $L_{\rm text}$ & 81.85 & 74.5 & 67.36 & 10.22 \\
    $R_{\rm img}$ & $388 \times 476$ & $467 \times 442$ & $402 \times 457$ & $2605 \times 2662$ \\
    \bottomrule
    \end{tabular}%
  \label{tab:datasets}%
\end{table}

\section{Notations}
\label{app:notations}

Table~\ref{tab:notations} summarizes the key mathematical notations used in this paper.
Additionally, Table~\ref{tab:datasets} provides the basic statistics and topological properties of the evaluated datasets, including graph scale (\#Nodes, \#Edges), category counts (\#Class), the homophily ratio ($H_{\text{hom}}$), and modality-specific metrics ($L_{\text{text}}$ and $R_{\text{img}}$).


\section{Proofs of Theoretical Analysis}
\label{sec:supra_proofs}

In this appendix, we provide complete proofs for the theoretical results stated in Section~\ref{sec:theory}. We use the notation established in Section~\ref{sec:pre}.

\subsection{Information-Theoretic Framework}
\label{app:mi_framework}

This section presents the information-theoretic definitions used in our analysis.

\begin{definition}[Semantic Signal-to-Noise Ratio]
\label{def:snr}
For a node $v$ and modality $m$, model the pre-trained feature as $\mathbf{x}_v^{(m)} = \mathbf{s}_v^{(m)} + \boldsymbol{\epsilon}_v^{(m)}$, where $\mathbf{s}_v^{(m)}$ is the true semantic signal and $\boldsymbol{\epsilon}_v^{(m)}$ is encoder noise with $\mathbb{E}[\|\boldsymbol{\epsilon}_v^{(m)}\|^2] = \sigma_\epsilon^2$. The \textbf{Semantic SNR} is:
\begin{equation}
    \text{SNR}_{int}^{(m)} = \frac{\|\mathbf{s}_v^{(m)}\|^2}{\sigma_\epsilon^2}.
\end{equation}
\end{definition}

\begin{definition}[Unique and Synergistic Mutual Information]
\label{def:mi_decompose}
Let $Y$ denote the target label. The \textbf{Total Predictive Information} $I(\mathbf{X}^{(t)}, \mathbf{X}^{(v)}; Y)$ decomposes as:
\begin{align}
    I_{\text{prior}} &= I_{\text{unique}}^{(t)} + I_{\text{unique}}^{(v)} + I_{\text{syn}}, \label{eq:mi_total} \\
    I_{\text{unique}}^{(m)} &= I(\mathbf{X}^{(m)}; Y), \quad I_{\text{syn}} = I(\mathbf{X}^{(t)}, \mathbf{X}^{(v)}; Y) - I_{\text{unique}}^{(t)} - I_{\text{unique}}^{(v)}, \label{eq:mi_syn}
\end{align}
where $I_{\text{unique}}^{(m)}$ is the predictive information from modality $m$ alone, and $I_{\text{syn}}$ is the \textbf{synergistic information}.
\end{definition}

\subsection{Proof of Theorem~\ref{thm:snr}: SNR Degradation under Mean-Aggregation}
\label{app:proof_thm1}

\textbf{Part (a): SNR Crossover and Degradation}

Consider a single modality $m$ processed by mean-aggregation (Eq~\ref{eq:mean_aggr}). For a node $v$, the pre-trained feature decomposes as $\mathbf{x}_v^{(m)} = \mathbf{s}_v^{(m)} + \boldsymbol{\epsilon}_v^{(m)}$, where $\mathbf{s}_v^{(m)}$ is the true semantic signal and $\boldsymbol{\epsilon}_v^{(m)}$ is encoder noise with $\mathbb{E}[\boldsymbol{\epsilon}_v^{(m)}] = \mathbf{0}$ and $\mathbb{E}[\|\boldsymbol{\epsilon}_v^{(m)}\|^2] = \sigma_\epsilon^2$.

After one layer of per-modality message passing with a self-loop weight $\alpha \in (0,1)$, the aggregated representation is:
\begin{equation}
    \mathbf{h}_v^{(m)} = \alpha \cdot \mathbf{x}_v^{(m)} + (1-\alpha) \cdot \frac{1}{|\mathcal{N}_v|} \sum_{u \in \mathcal{N}_v} \mathbf{x}_u^{(m)}.
\end{equation}

First, we decompose the neighborhood average $\bar{\mathbf{x}}_{\mathcal{N}}^{(m)} = \frac{1}{|\mathcal{N}_v|} \sum_{u \in \mathcal{N}_v} \mathbf{x}_u^{(m)}$ into a homophily-aligned signal and residual topological noise:
\begin{equation}
    \bar{\mathbf{x}}_{\mathcal{N}}^{(m)} = \beta \cdot \mathbf{s}_v^{(m)} + \boldsymbol{\xi}_v^{(m)},
\end{equation}

where $\beta \in [0,1]$ captures the average semantic alignment between $v$ and its neighbors, and $\mathbb{E}[\|\boldsymbol{\xi}_v^{(m)}\|^2] = \sigma_{\mathcal{N}}^2$. Substituting this into the aggregation yields:
\begin{equation}
    \mathbf{h}_v^{(m)} = \underbrace{(\alpha + (1-\alpha)\beta) \cdot \mathbf{s}_v^{(m)}}_{\text{effective signal}}
    + \underbrace{\alpha \cdot \boldsymbol{\epsilon}_v^{(m)} + (1-\alpha) \cdot \boldsymbol{\xi}_v^{(m)}}_{\text{effective noise}}.
\end{equation}

To compute the expected post-aggregation SNR, we formally assume that the intrinsic semantic signal $\mathbf{s}_v^{(m)}$, the encoder noise $\boldsymbol{\epsilon}_v^{(m)}$, and the induced neighborhood noise $\boldsymbol{\xi}_v^{(m)}$ are mutually independent. Consequently, their cross-correlation terms vanish in expectation, allowing the variances to add linearly. The post-aggregation SNR is then given by:
\begin{equation}
\label{eq:snr_post_app}
    \text{SNR}_{post}^{(m)} = \frac{(\alpha + (1-\alpha)\beta)^2 \|\mathbf{s}_v^{(m)}\|^2}{\alpha^2 \sigma_\epsilon^2 + (1-\alpha)^2 \sigma_{\mathcal{N}}^2}.
\end{equation}

Next, we compare $\text{SNR}_{post}^{(m)}$ with the intrinsic SNR ($\text{SNR}_{int}^{(m)} = \|\mathbf{s}_v^{(m)}\|^2 / \sigma_\epsilon^2$) to determine the crossover condition:
\begin{align}
    \text{SNR}_{post}^{(m)} - \text{SNR}_{int}^{(m)}
    &= \|\mathbf{s}\|^2 \cdot \frac{(\alpha + (1-\alpha)\beta)^2 \sigma_\epsilon^2 - \bigl[\alpha^2 \sigma_\epsilon^2 + (1-\alpha)^2 \sigma_{\mathcal{N}}^2\bigr]}{\sigma_\epsilon^2\bigl(\alpha^2 \sigma_\epsilon^2 + (1-\alpha)^2 \sigma_{\mathcal{N}}^2\bigr)}.
\end{align}
The sign of this difference is determined by the numerator. We define the constant $A$ as:
\begin{equation}
    A \coloneqq (\alpha + (1-\alpha)\beta)^2 - \alpha^2 = (1-\alpha)\beta \cdot \bigl[2\alpha + (1-\alpha)\beta\bigr] > 0.
\end{equation}
Setting $A \cdot \sigma_\epsilon^2 - (1-\alpha)^2 \sigma_{\mathcal{N}}^2 = 0$ provides the critical noise threshold:
\begin{equation}
\label{eq:tau_app}
    \tau(\beta) = \frac{(1-\alpha)^2 \sigma_{\mathcal{N}}^2}{A} = \frac{1-\alpha}{\beta\bigl(2\alpha + (1-\alpha)\beta\bigr)}\;\sigma_{\mathcal{N}}^2.
\end{equation}

Since $A > 0$ for any $\beta > 0$, the threshold $\tau(\beta)$ is strictly positive. Therefore, the crossover condition can be precisely stated as:
\begin{equation}
\label{eq:crossover_app}
    \text{SNR}_{post}^{(m)} \gtrless \text{SNR}_{int}^{(m)} \quad\Longleftrightarrow\quad \sigma_\epsilon^2 \gtrless \tau(\beta).
\end{equation}
This establishes two distinct regimes: (1) for \textit{weak features} ($\sigma_\epsilon^2 > \tau(\beta)$), encoder noise dominates, and GNN aggregation yields an SNR improvement; (2) for \textit{strong features} ($\sigma_\epsilon^2 < \tau(\beta)$), encoder noise is already minimized, and the fixed structural cost $(1-\alpha)^2\sigma_{\mathcal{N}}^2$ introduced by aggregation inevitably degrades the SNR.

Finally, examining the limiting behavior as feature quality approaches perfection ($\sigma_\epsilon^2 \to 0$):
\begin{equation}
    \frac{\text{SNR}_{post}^{(m)}}{\text{SNR}_{int}^{(m)}} \;\xrightarrow{\sigma_\epsilon^2 \to 0}\; 0.
\end{equation}
This confirms that in the limit of high-confidence semantic priors, mandatory topological aggregation destroys the relative SNR advantage, a representational collapse analogous to the asymptotic exponential loss of expressive power in deep GNNs~\cite{oono2019graph,chen2020measuring}. \qed

\textbf{Part (b): Information-Theoretic Upper Bound}

We proceed to prove that mean-aggregation imposes a strict information-theoretic ceiling. 

Consider the Markov chain $Y \to (\mathbf{X}^{(t)}, \mathbf{X}^{(v)}) \to \mathbf{h}^{(\text{final})}$. By applying the Data Processing Inequality (DPI), we establish that the final predictive information is bounded by the initial prior information:
\begin{equation}
    I(\mathbf{h}^{(\text{final})}; Y) \leq I(\mathbf{X}^{(t)}, \mathbf{X}^{(v)}; Y) = I_{\text{prior}}.
\end{equation}

We define the topological information loss for each modality $m$ as $\Delta_{\text{topo}}^{(m)} \coloneqq I(\hat{\mathbf{x}}^{(m)}; Y) - I(\hat{\mathbf{h}}^{(m)}; Y)$, which is inherently non-negative due to the DPI applied to the local aggregation pathway $Y \to \hat{\mathbf{x}}^{(m)} \to \hat{\mathbf{h}}^{(m)}$. 

Under a Gaussian approximation, the mutual information $I(\hat{\mathbf{h}}^{(m)}; Y)$ is monotonically increasing with respect to the post-aggregation SNR. As established in Part (a), when $\sigma_\epsilon^2 < \tau(\beta)$, the post-aggregation SNR is strictly degraded ($\text{SNR}_{post}^{(m)} < \text{SNR}_{int}^{(m)}$). This directly implies that $I(\hat{\mathbf{h}}^{(m)}; Y) < I(\hat{\mathbf{x}}^{(m)}; Y)$, ensuring $\Delta_{\text{topo}}^{(m)} > 0$.

Consequently, the total information loss $\Delta_{\text{topo}} = \sum_{m} \Delta_{\text{topo}}^{(m)}$ is strictly positive in the high-confidence regime. As feature quality improves (i.e., $\sigma_\epsilon^2 / \tau(\beta)$ decreases), the SNR degradation worsens, causing $\Delta_{\text{topo}}^{(m)}$ to grow. Furthermore, because the intrinsic geometric structure of individual modalities is distorted by topological noise, their synergistic mutual information $I_{\text{syn}}$ is also inherently compromised, compounding the total information loss.
This formalizes the semantic-topological trade-off:
\begin{equation}
    I_{\text{final}} \leq I_{\text{prior}} - \Delta_{\text{topo}}, \quad \text{where } \Delta_{\text{topo}} > 0 \text{ when } \sigma_\epsilon^2 < \tau(\beta).
\end{equation}
\qed

\subsection{Proof of Proposition~\ref{prop:starvation}: Gradient Starvation under Modality Asymmetry}
\label{app:proof_thm2}

We prove that two mechanistically independent phenomena---architectural gradient dilution and modality competition---concurrently cause gradient starvation for the weaker modality encoder. To maintain notational clarity, we denote an arbitrary node as $i$, and the text and visual modalities as $t$ and $v$, respectively, with corresponding encoders $f_t$ and $f_v$.

\textbf{Part (a): Architectural Gradient Dilution}

First, we analyze the architectural gradient attenuation through topological aggregation. Consider a single mean-aggregation layer for a specific node $i$: $\mathbf{h}_i^{(\ell)} = \alpha \cdot \mathbf{h}_i^{(\ell-1)} + (1-\alpha) \cdot \bar{\mathbf{h}}_{\mathcal{N}_i}^{(\ell-1)}$. The Jacobian of node $i$'s output with respect to its own representation is given by:
\begin{equation}
    \frac{\partial \mathbf{h}_i^{(\ell)}}{\partial \mathbf{h}_i^{(\ell-1)}} = \alpha \cdot \mathbf{I} + (1-\alpha) \cdot \frac{\partial \bar{\mathbf{h}}_{\mathcal{N}_i}^{(\ell-1)}}{\partial \mathbf{h}_i^{(\ell-1)}}.
\end{equation}
While the theoretical upper bound of the full spatial Jacobian norm could be 1, the structural gradient flowing directly through the \textit{ego-node pathway}---which carries the node's uncorrupted intrinsic semantic signal---is explicitly determined by the self-loop term $\alpha \cdot \mathbf{I}$. Assuming the gradients from neighbors act as zero-mean topological noise that cancel out in expectation across diverse neighborhoods, the effective expected gradient magnitude for the ego-pathway is dominated by $\alpha$.

Applying the chain rule across $L$ layers specifically for this ego-node pathway, we obtain:
\begin{equation}
    \mathbb{E}\left[ \frac{\partial \mathbf{h}_i^{(L)}}{\partial \mathbf{h}_i^{(0)}} \right] \approx \prod_{\ell=1}^{L} (\alpha \cdot \mathbf{I}) = \alpha^L \cdot \mathbf{I}.
\end{equation}
Since $\alpha \in (0,1)$, the direct structural gradient contribution exhibits an exponential decay $\mathcal{O}(\alpha^L)$ with respect to the network depth $L$, establishing a severe architectural bottleneck. \qed

\textbf{Part (b): Modality Competition via SNR Gap}

Next, we derive the modality competition mechanism arising from the SNR asymmetry. Consider a linear multimodal prediction head for node $i$:
\begin{equation}
\label{eq:linear_pred_app}
    \hat{y}_i = \mathbf{w}_t^\top \mathbf{h}_i^{(t)} + \mathbf{w}_v^\top \mathbf{h}_i^{(v)},
\end{equation}
where $\mathbf{h}_i^{(m)} = f_m(\mathbf{x}_i^{(m)})$ is the encoded representation. Under the mean squared error loss $\mathcal{L}_i = \frac{1}{2}(\hat{y}_i - y_i)^2$, the gradients with respect to the encoders are:
\begin{align}
    \nabla_{f_t} \mathcal{L}_i &= (\hat{y}_i - y_i) \cdot \mathbf{w}_t \cdot \frac{\partial \mathbf{h}_i^{(t)}}{\partial f_t}, \label{eq:grad_t_app} \\
    \nabla_{f_v} \mathcal{L}_i &= (\hat{y}_i - y_i) \cdot \mathbf{w}_v \cdot \frac{\partial \mathbf{h}_i^{(v)}}{\partial f_v}. \label{eq:grad_v_app}
\end{align}

Because the text modality possesses a significantly higher intrinsic SNR ($\text{SNR}_{int}^{(t)} \gg \text{SNR}_{int}^{(v)}$), its corresponding branch rapidly reduces the prediction error during the early phase of joint training. As the weights $\mathbf{w}_t$ converge, the joint residual $r_i(t) = \hat{y}_i(t) - y_i$ is effectively suppressed by the text branch:
\begin{equation}
\label{eq:residual_decay}
    |r_i(t)| \;\leq\; |y_i - \mathbf{w}_t^\top \mathbf{h}_i^{(t)}| + |\mathbf{w}_v^\top \mathbf{h}_i^{(v)}| \;\xrightarrow{t \to \infty}\; \mathcal{O}(\sigma_{\epsilon,v}).
\end{equation}

Consequently, the gradient for the weaker visual encoder (Eq. \ref{eq:grad_v_app}) is directly proportional to this vanishing residual:
\begin{equation}
\label{eq:grad_v_residual}
    \|\nabla_{f_v} \mathcal{L}_i\| = |r_i(t)| \cdot \|\mathbf{w}_v\| \cdot \left\|\frac{\partial \mathbf{h}_i^{(v)}}{\partial f_v}\right\|.
\end{equation}
As $r_i(t) \to 0$, the visual gradient vanishes ($\|\nabla_{f_v} \mathcal{L}_i\| \to 0$) even if the visual representation is sub-optimal. This dynamics manifests as a form of {Shortcut Learning}~\cite{OGMGE}, where the optimizer discovers that the text branch is sufficient and effectively treats the visual branch as redundant. Consequently, the gradient ratio $R(t) = \|\nabla_{f_v} \mathcal{L}_i(t)\| / \|\nabla_{f_t} \mathcal{L}_i(t)\|$ rapidly approaches zero. 

Crucially, this modality competition is an emergent property of optimization dynamics under SNR asymmetry, rendering it mechanistically independent of the deterministic architectural dilution analyzed in Part (a). \qed

\textbf{Part (c): Combined Effect}

Finally, by unifying the architectural dilution (Part a) and modality competition (Part b), the expected gradient norm for the weaker visual encoder $f_v$ is strictly bounded by:
\begin{equation}
    \mathbb{E}\left[\|\nabla_{f_v}\mathcal{L}_{task}\|\right] \;\leq\; \underbrace{\alpha^L}_{\substack{\text{architectural} \\ \text{(Part a)}}} \cdot \underbrace{|r_i(t)| \cdot \|\mathbf{w}_v\| \cdot \left\|\frac{\partial \mathbf{h}_i^{(v)}}{\partial f_v}\right\|}_{\substack{\text{competition} \\ \text{(Part b)} \to 0}}.
\end{equation}
Since $\alpha^L < 1$ and $|r_i(t)| \to 0$, their product rigorously explains the vanishing gradients for the visual encoder, formalizing the dual-bottleneck pathology of Gradient Starvation. \qed

\subsection{Analysis of Independent Aggregation Architectures}
\label{app:proof_cor_ip}

This section proves that architectures based on Independent Aggregation remain vulnerable to the pathologies identified in Theorems~\ref{thm:snr} and Proposition~\ref{prop:starvation}.

\begin{corollary}[Vulnerability of Independent Aggregation]
\label{cor:independent_failure}
Architectures that process each modality through separate GNN branches before fusion remain susceptible to SNR degradation and gradient starvation. Specifically, independent aggregation cannot achieve the optimal gradient throughput of $\mathcal{O}(1)$ afforded by the decoupled architecture of \model.
\end{corollary}

\subsubsection*{Part (a): Intra-modal Topological Degradation}

In Independent Aggregation, modality $m$ is processed through its own GCN branch:
\begin{equation}
    \mathbf{Z}_v^{(m)} = \text{GCN}_m\!\left(\mathbf{A}, \mathbf{X}^{(m)}\right),
\end{equation}
where each layer applies mean-aggregation: $\mathbf{h}_v^{(m,\ell)} = \alpha_m \cdot \mathbf{h}_v^{(m,\ell-1)} + (1-\alpha_m) \cdot \bar{\mathbf{h}}_{\mathcal{N}}^{(m,\ell-1)}$.

For the text modality with $\sigma_{\epsilon,t}^2 < \tau_t(\beta_t)$, the post-aggregation SNR within the text branch satisfies:
\begin{equation}
    \text{SNR}_{post}^{(t,\text{Ind.})} \;<\; \text{SNR}_{int}^{(t)},
\end{equation}
by Theorem~\ref{thm:snr}(a). The branch-level separation prevents cross-modal contamination, so:
\begin{equation}
    \text{SNR}_{post}^{(t,\text{Joint})} \;<\; \text{SNR}_{post}^{(t,\text{Ind.})} \;<\; \text{SNR}_{int}^{(t)}.
\end{equation}
Independent Aggregation is better than Joint Aggregation for the strong modality, but still worse than no aggregation at all. \qed

\subsubsection*{Part (b): Persistent Gradient Starvation}

Despite independent GCN branches, Independent Aggregation converges to a shared classifier:
\begin{equation}
    \hat{y} = \text{Head}\!\left([\mathbf{Z}^{(t)} \| \mathbf{Z}^{(v)}]\right) = \mathbf{w}_t^\top \mathbf{Z}^{(t)} + \mathbf{w}_v^\top \mathbf{Z}^{(v)},
\end{equation}
supervised by a single task loss $\mathcal{L}_{task}$. By Proposition~\ref{prop:starvation}(b), the SNR gap $\text{SNR}^{(t)} \gg \text{SNR}^{(v)}$ causes the text branch to rapidly converge, suppressing the joint residual $|r(t)| \to 0$. Since both branches' gradients are proportional to $r(t)$, the visual branch's gradient vanishes:
\begin{equation}
    \|\nabla_{f_v} \mathcal{L}_{task}\| = |r(t)| \cdot \|\mathbf{w}_v\| \cdot \left\|\frac{\partial \mathbf{Z}^{(v)}}{\partial \mathbf{H}^{(v)}} \cdot \frac{\partial \mathbf{H}^{(v)}}{\partial f_v}\right\| \;\to\; 0.
\end{equation}

Additionally, each branch's gradient must pass through its own GCN's Jacobian, introducing $\alpha_v^{L_v}$ decay (Proposition~\ref{prop:starvation}(a)):
\begin{equation}
    \|\nabla_{f_v}\mathcal{L}_{\text{task}}\| \;\leq\; \underbrace{\alpha_v^{L_v}}_{\text{architectural}} \cdot \underbrace{|r(t)| \cdot \|\mathbf{w}_v\| \cdot \left\|\tfrac{\partial \mathbf{H}^{(v)}}{\partial f_v}\right\|}_{\to\, 0\text{ (competition)}}.
\end{equation}

\textbf{Key difference from SUPRA:} Independent Aggregation provides no topology-free gradient pathway. In SUPRA, the Unique Stream $\mathbf{Z}^{(U_m)} = f_m(\mathbf{X}^{(m)})$ bypasses the GNN entirely, providing an $\mathcal{O}(1)$ gradient path. Even if one were to add an auxiliary loss to Independent Aggregation's branch outputs, the auxiliary gradient would still pass through the GCN Jacobian ($\mathcal{O}(\alpha_v^{L_v})$), not $\mathcal{O}(1)$. \qed

\section{Limitations}
\label{app:limitations_impact}

\subsection{Limitations}
While \model effectively resolves the \textit{aggregation dilemma} through a decoupled dual-pathway architecture, we acknowledge several limitations that offer opportunities for future research:

\begin{itemize}[leftmargin=*]
    \item \textbf{Scalability to Massive-Scale Graphs:} Although we have demonstrated significant efficiency gains (3.5$\times$ memory reduction), our current evaluation is focused on representative medium-to-large benchmarks. Further validation is required to assess \model's generalization and performance on massive-scale graphs exceeding millions of nodes and edges. Exploring its performance in industrial-scale distributed environments remains a key future direction.
    
    \item \textbf{Adaptive Optimization Objectives:} In our current implementation, the auxiliary supervision weight $\lambda_{aux}$ is a hyperparameter that requires manual tuning. While our sensitivity analysis shows it is relatively robust, a more principled approach—such as meta-learning or uncertainty-based weighting schemes—could enable \model to adaptively adjust the optimization balance between specificity and synergy streams.
    
    \item  \textbf{Adaptive Inference Mechanism:} Currently, \model utilizes a straightforward pooling strategy for joint prediction at inference time. While this is efficient, there is significant potential for designing more adaptive inference mechanisms. For instance, allowing each node to dynamically weigh the contributions of the specificity and synergy streams based on its local neighborhood density or semantic confidence could further enhance the model's predictive precision.
\end{itemize}

\end{document}